%% file: main.tex
\PassOptionsToPackage{table}{xcolor}
\documentclass[manuscript,authorversion,nonacm]{acmart}
\usepackage[english]{babel}
\usepackage{adjustbox}
\usepackage{caption}
\usepackage{subcaption}
\usepackage{commath}
\usepackage{algorithm2e}
\usepackage{svg}
\usepackage{makecell}
\usepackage{graphicx}
\usepackage{soul}
\usepackage{pifont}
\usepackage{xspace}
\usepackage{color, colortbl}
\usepackage{float}
\usepackage{booktabs, multirow} 
\usepackage{amsmath}
\usepackage{dsfont}
\usepackage[skip=0.5\baselineskip]{caption}

\usepackage{mwe}
\usepackage{enumitem}
\usepackage{amsthm}
\usepackage{caption}
\usepackage{bbm}
\usepackage{tikz}
\usepackage{xcolor}

\usepackage{pgfplots}
\pgfplotsset{compat=1.17}
\definecolor{clrLR}{RGB}{175, 213, 79}
\definecolor{clrDT}{RGB}{225, 160, 79}
\definecolor{clrSVM}{RGB}{225, 0, 79}
\definecolor{clrLGB}{RGB}{207, 52, 118}
\definecolor{clrXGB}{RGB}{32, 137, 227}
\definecolor{clRF}{RGB}{254, 111, 0}
\definecolor{clMLP}{RGB}{13, 6, 134}
\definecolor{clLFERM}{RGB}{134, 197, 189}
\definecolor{clADV}{RGB}{129, 12, 134}
\definecolor{clZafar}{RGB}{252, 119, 147}

\theoremstyle{definition}
\newtheorem{definition}{Definition}

\AtBeginDocument{%
  \providecommand\BibTeX{{%
    \normalfont B\kern-0.5em{\scshape i\kern-0.25em b}\kern-0.8em\TeX}}}

\begin{document}
\title{Counterfactual Fair Opportunity: Measuring Decision Model Fairness with Counterfactual Reasoning}

 \author{Giandomenico Cornacchia}
 \email{giandomenico.cornacchia@poliba.it}
 \orcid{0000-0001-5448-9970}
 \affiliation{%
  \institution{Polytechnic University of Bari}
  \streetaddress{Via Orabona, 4}
  \city{Bari}
  \country{Italy}
  \postcode{70125}
 }
 \author{Vito Walter Anelli}
 \email{vitowalter.anelli@poliba.it}
 \orcid{0000-0002-5567-4307}
 \affiliation{%
  \institution{Polytechnic University of Bari}
  \streetaddress{Via Orabona, 4}
  \city{Bari}
  \country{Italy}
  \postcode{70125}
 }
 \author{Fedelucio Narducci}
 \email{fedelucio.narducci@poliba.it}
 \orcid{0000-0002-9255-3256}
 \affiliation{%
  \institution{Polytechnic University of Bari}
  \streetaddress{Via Orabona, 4}
  \city{Bari}
  \country{Italy}
  \postcode{70125}
 }
 \author{Azzurra Ragone}
 \affiliation{%
  \institution{University of Bari}
  \streetaddress{Piazza Umberto I, 1}
  \city{Bari}
  \country{Italy}
  \postcode{70121}}
 \email{azzurra.ragone@uniba.it}
 \orcid{0000-0002-3537-7663}
 
 \author{Eugenio Di Sciascio}
 \email{eugenio.disciascio@poliba.it}
 \orcid{0000-0002-5484-9945}
 \affiliation{%
  \institution{Polytechnic University of Bari}
  \streetaddress{Via Orabona, 4}
  \city{Bari}
  \country{Italy}
  \postcode{70125}
 }
\renewcommand{\shortauthors}{G. Cornacchia, et~al.}

\begin{abstract}
\input{Sections/0_abstract.tex}
\end{abstract}





\maketitle

\section{Introduction}\label{sec:introduction}
\input{Sections/1_Introduction.tex}
\section{Related Work}\label{sec:bibliography}
\input{Sections/2_Related.tex}
\section{Preliminaries}\label{sec:preliminaries}
\input{Sections/3_Preliminaries.tex}

\section{Methodology}\label{sec:approach}
\input{Sections/4_Methodology.tex}
\section{Experimental Evaluation}\label{sec:experiments}
\input{Sections/5_Experiments.tex}
\section{Results}\label{sec:results}
\input{Sections/6_Results.tex}

\section{Conclusion}\label{sec:conclusion}
\input{Sections/7_Conlusion.tex}


\bibliographystyle{ACM-Reference-Format}
\bibliography{main}

\pagebreak
\appendix
\section{Additional Experiments}\label{sec:appendix}
\input{tables/CFstatistics.tex}
\input{tables/CFlipsGeneticNegOtherSensitive.tex}
\input{tables/CFlipsKDtreeNegOtherSensitive.tex}
\input{tables/DeltaNDCCFRerank.tex}
\input{tables/DeltaNDCCFRerankOtherSensitive.tex}
\end{document}

%% file: Sections/0_abstract.tex
The increasing application of Artificial Intelligence and Machine Learning models poses potential risks of unfair behavior and, in light of recent regulations, has attracted the attention of the research community.
Several researchers focused on seeking new fairness definitions or developing approaches to identify biased predictions.
However, none try to exploit the counterfactual space to this aim.
In that direction, the methodology proposed in this work aims to unveil unfair model behaviors using counterfactual reasoning in the case of fairness under unawareness setting.
 A counterfactual version of \textit{equal opportunity} named \textit{counterfactual fair opportunity} is defined and
two novel metrics that analyze the sensitive information of counterfactual samples are introduced. 
Experimental results on three different datasets show the efficacy of our methodologies and our metrics, disclosing the unfair behavior of classic machine learning and debiasing models.



%% file: Sections/1_Introduction.tex
Artificial Intelligence (AI) systems are increasingly pervasive in our society and are also often exploited for taking life-changing decisions, like e.g. loans, job offers, and health care access. 
However, in these domains one should be aware of the potential risk of \textit{discrimination}\footnote{\url{https://dictionary.cambridge.org/dictionary/english/discrimination}} of groups or individuals.

In the fintech industry, for instance,  \textit{online instant lending platforms} use machine learning tools to analyze available consumer credit data to make faster credit decisions.
Nevertheless, in the financial sector, the choice to grant or deny credit is regulated by rigorous and thorough regulatory compliance criteria referring primarily to human decisions (e.g., Equal Credit Opportunity Act\footnote{\url{https://www.ftc.gov/enforcement/statutes/equal-credit-opportunity-act}} 
and Consumer Credit Directive for EU Community\footnote{\url{https://eur-lex.europa.eu/legal-content/en/TXT/?uri=CELEX:32008L0048}}). 

Nonetheless, when AI replaces human decisions, like in the case of instant lending, there is a risk of revealing a loophole in existing liability identification laws.
Several national and international organizations have released guidelines, norms, and principles to prevent the irresponsible usage of AI, e.g., EU Commission with ``Ethics Guidelines for Trustworthy AI'' and more recently ``The Proposal for Harmonized Rule on AI'', United States with the ``AI Bill of Rights''~\footnote{\url{https://www.whitehouse.gov/ostp/ai-bill-of-rights/}}, or the expert group on ``AI in Society'' of the Organisation for Economic Co-operation and Development~(OECD).

Although scientists train their models without explicit discriminating intents, deploying AI systems without considering ethical concerns may lead to discrimination~\citep{DBLP:conf/kdd/Corbett-DaviesP17}. Even more problematic is figuring out which type of discrimination is being implemented~\citep{Castelnovo2021TheZO}.
In the last years, a wide range of definitions has been proposed for fairness, however, the application of a fairness concept might go against another. This makes the choice of a criterion ethically challenging.
This work exploits an approach for detecting decision biases in a counterfactual space, proposing a new counterfactual fairness definition: the \textit{Counterfactual Fair Opportunity}.
The problem addressed in this paper is to identify whether a model is biased even though it does not use sensitive features, as requested by regulations in some domains (e.g., finance or healthcare).
Accordingly, we proposed an approach based on counterfactual reasoning to detect biases in \textit{fairness under unawareness} setting.


We introduce two novel fairness metrics, \textit{Counterfactual Flips} (CFlips) and \textit{normilized Discounted Cumulative Counterfactual Fairness} (nDCCF), able to identify the discriminatory behavior of a \textbf{Decision Maker} exploiting a \textbf{Countrerfactual Generator} and an oracle, i.e., the \textbf{Sensitive-Feature Classifier}. Our auditing methodology aims to be an effective tool for quantifying the discriminatory behavior of any ML model since it works in a black-box setting. 

The remainder of the paper is organized as follows: Section~\ref{sec:bibliography} provides the background of the most important fairness desiderata and an overview of the most relevant research in the fields of fairness and counterfactual reasoning. Section~\ref{sec:preliminaries} introduces the notation, and Section~\ref{sec:approach} depicts the methodology and the metrics based on our fairness definition. Section~\ref{sec:experiments} describes the experimental evaluation, whereas the results are discussed in Section~\ref{sec:results}. The conclusions are drawn in Section~\ref{sec:conclusion}. 

%% file: Sections/2_Related.tex
Fairness is a well-studied topic with a considerable body of knowledge to draw from~\citep{DBLP:journals/ipm/AshokanH21, DBLP:conf/cikm/ZhuHC18,DBLP:conf/kdd/PedreschiRT08,DBLP:conf/inns/OnetoC19}. 
The scientific community has drawn up a wide range of fairness definitions that are derived from specific legal, philosophical, or mathematical applications~\cite{verma2018fairness}. To name a few, \textit{Fairness under Unawareness}, \textit{Disparate Impact}, \textit{Independence}, \textit{Separation}, and \textit{Counterfactual Fairness}. 
An increasing number of algorithms have been proposed to implement a specific concept of fairness.
Unfortunately, if we make an algorithm fair on one measure, it could become unfair on another since the most often used criteria for fairness are frequently in conflict with each other~\citep{DBLP:conf/innovations/KleinbergMR17F}.
Generally, fairness definitions refer to people characterized by sensitive information (e.g., gender, age, race) and separable into privileged and unprivileged groups, within which disparity metrics are measured (i.e., \textit{group-fairness}).

\citet{DBLP:conf/innovations/DworkHPRZ12} introduced the concept of \textit{individual-fairness} according to which similar individuals should be treated similarly. 
Specifically, from a geometrical perspective that considers how a decision model works, 
similar individuals should receive similar outcomes. 
On that concept lies the basis for the definition of \textit{Counterfactual Fairness}~\citep{DBLP:conf/nips/KusnerLRS17}: `\textit{a predictor can be considered counterfactually fair if its result does not change between individuals with the same characteristics but different sensitive information}'.
However, this definition requires to know sensitive features that are strictly forbidden in particular domains (e.g., finance, health care). The only setting that could be considered coherent with the law is the \textit{fairness under unawareness} since no sensitive information is used directly for the model training and for making decisions. Indeed, \textit{fairness under unawareness} states that "\textit{an algorithm is fair as long as any protected attributes are not explicitly used in the decision-making process}"~\cite{mehrabi2021survey}.

Unfortunately, removing sensitive information does not ensure the decision maker is fair due to the existence of \textit{proxy features}~\cite{CORNACCHIA2023103224,DBLP:conf/kdd/Corbett-DaviesP17, doi:10.1126/sciadv.aao5580, doi:10.1126/science.187.4175.398}.
In this regard, our model overcomes this limitation by being able to detect bias in the case of \textit{fairness under unawareness}~\citep{DBLP:conf/fat/ChenKMSU19} through the use of counterfactual reasoning~\citep{DBLP:conf/ecai/Pearl94}. A new fairness definition and two fairness metrics are proposed. The model works in a counterfactual space for detecting and quantifying discriminatory biases. The metrics base their definition on the \textit{group-fairness}, but their computation is based on the \textit{`counterfactual' individual-fairness}.
\input{groupplot/grouppedTSNE}

%% file: groupplot/grouppedTSNE.tex
\begin{figure*}[!t]
\begin{subfigure}{.245\textwidth}
\includegraphics[width=\textwidth,height=1\textwidth]{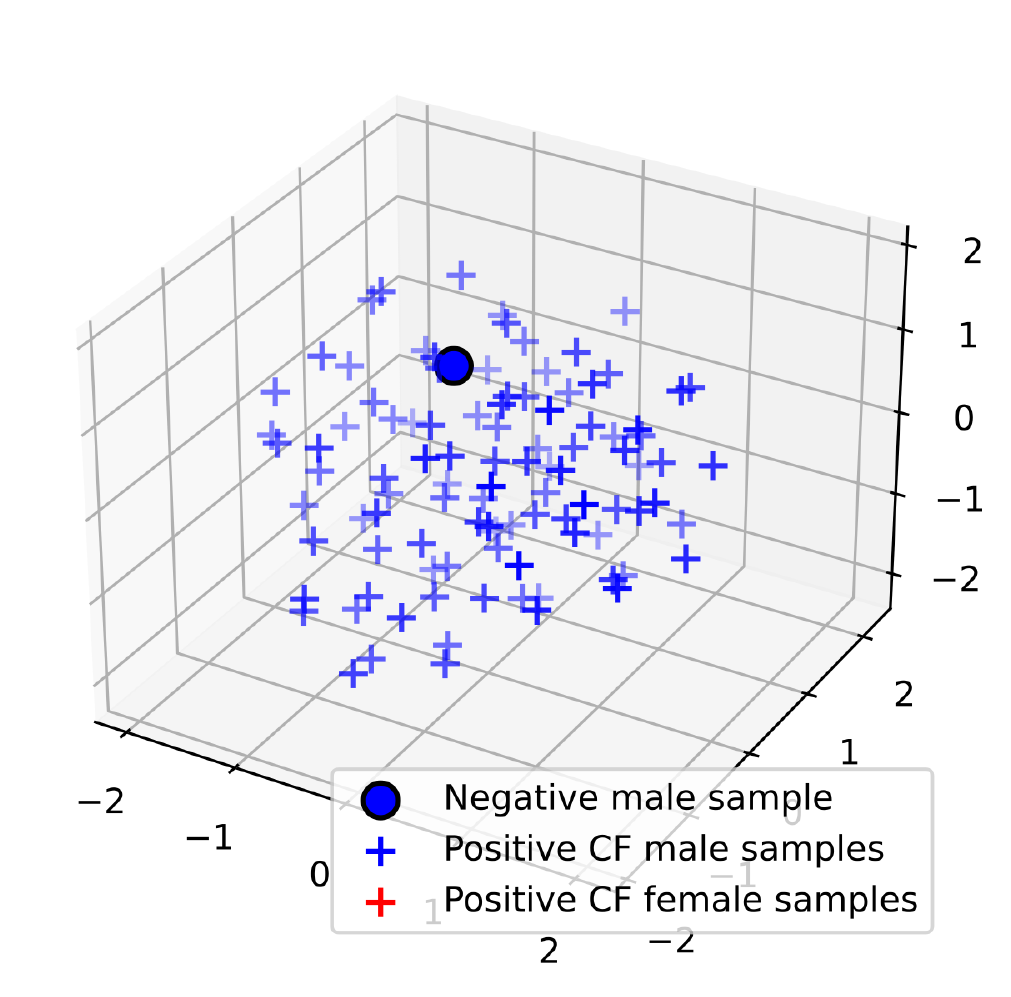}
\caption{male on Classic ML model}\label{fig:XGBmale}
\end{subfigure}
\begin{subfigure}{.245\textwidth}
\includegraphics[width=\textwidth,height=1\textwidth]{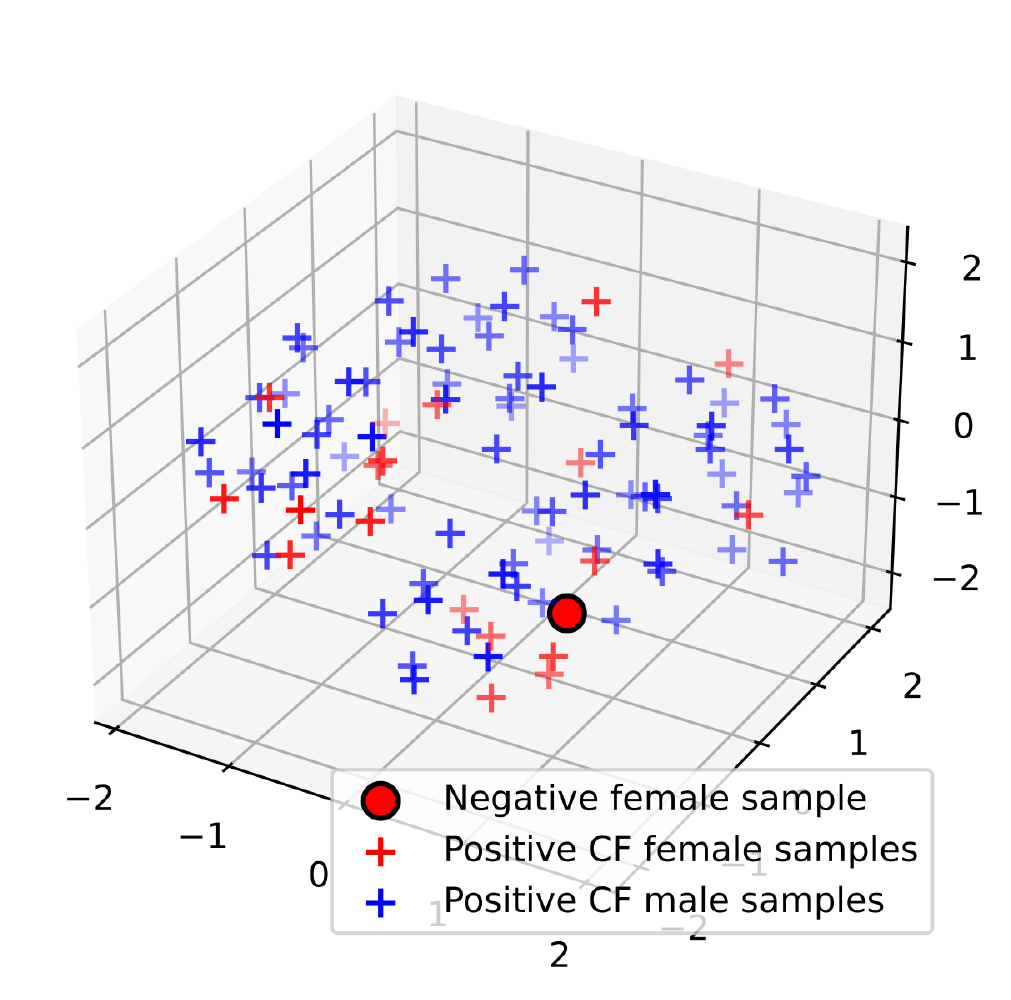}
\caption{female on Classic ML model}\label{fig:XGBfemale}
\end{subfigure}
\begin{subfigure}{.245\textwidth}
\includegraphics[width=\textwidth,height=1\textwidth]{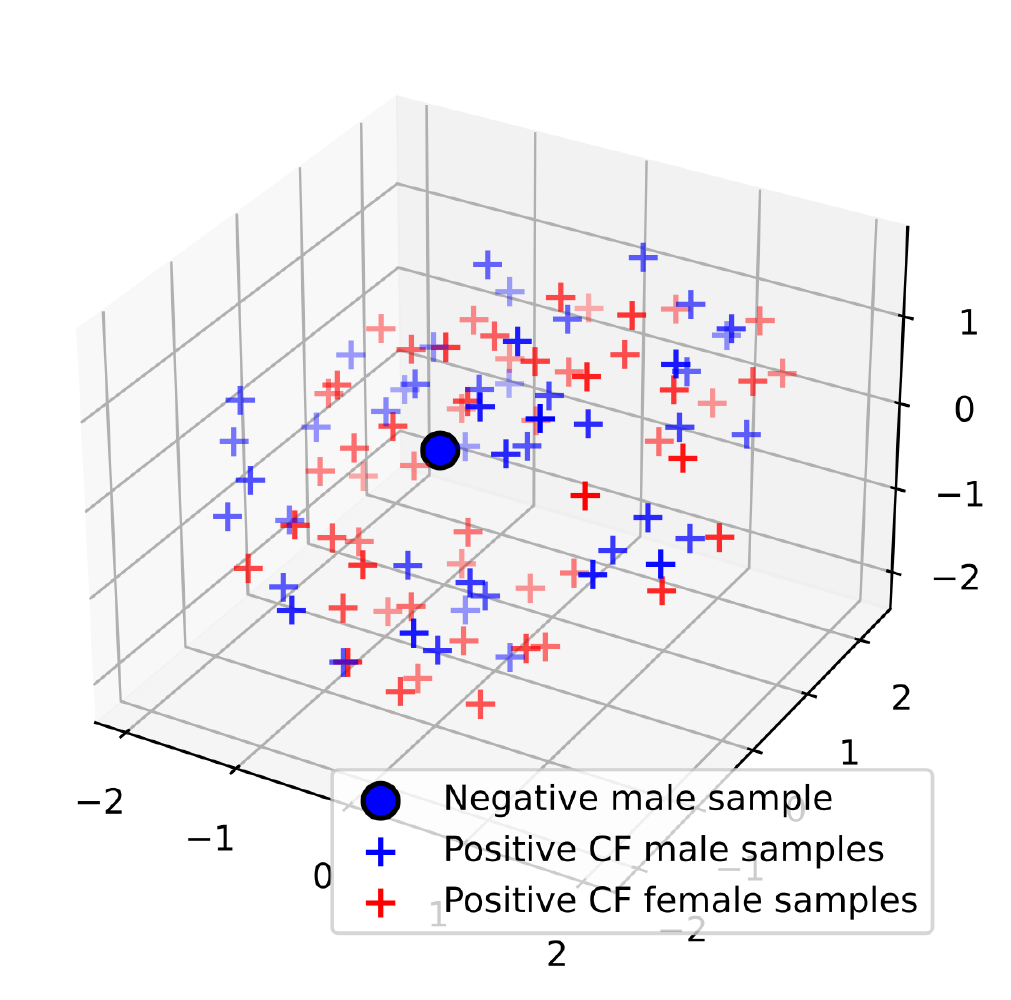}
\caption{male on Debiasing model}\label{fig:AdvDebMale}
\end{subfigure}
\begin{subfigure}{.245\textwidth}
\includegraphics[width=\textwidth,height=1\textwidth]{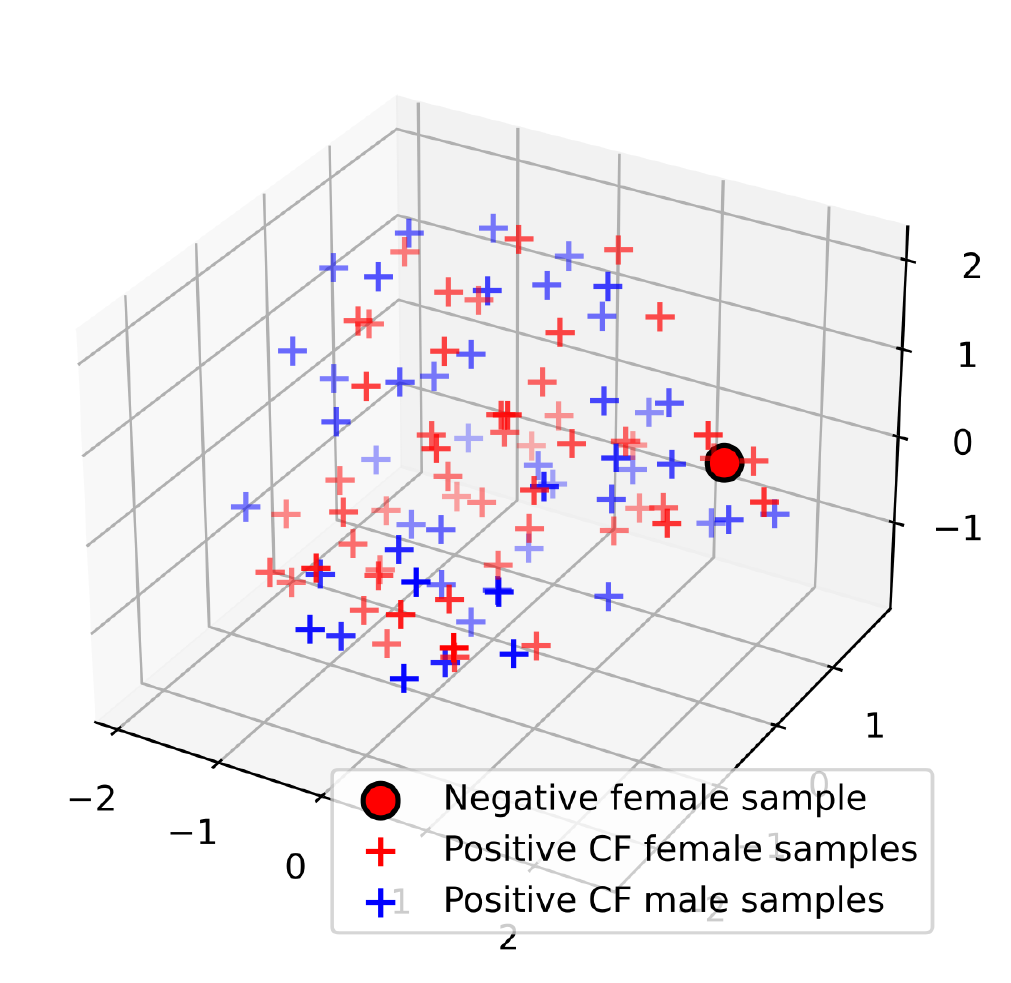}
\caption{female on Debiasing model}\label{fig:AdvDebFemale}
\end{subfigure}
\caption{Adult t-SNE visualizations of a random male (a-c) and female (b-d) sample with a negative outcome and their CF samples with a positive outcome, respectively, for a Classic ML model (i.e. XGB) and a Debiasing model (i.e. Adversarial~Debiasing).}
\label{fig:totalFigures}
\end{figure*}

%% file: Sections/3_Preliminaries.tex
This section introduces the notation used in the rest of the paper.

\noindent\textbf{Data points.} We assume the dataset $\mathcal{D}$ is an $m$-dimensional space containing $n$ non-sensitive features, $l$ sensitive features, and a target attribute. In other words, we have $\mathcal{D} \subseteq \mathbb{R}^{m}$, with $m = n+l+1$\footnote{Without loss of generality, we assume that categorical features can always be transformed into features in $\mathbb{R}$ via one-hot-encoding.}. A data point $d\in\mathcal{D}$ is then represented as $d=\langle \mathbf{x}, \mathbf{s}, y \rangle$, with $\mathbf{x} = \langle x_{1}, x_{2},..., x_{n}\rangle$ representing the sub-vector of non-sensitive features, $\mathbf{s} = \langle s_{1}, s_{2},..., s_{l}\rangle$ the sub-vector of sensitive features and $y$ being a binary target feature.
Given a vector of sensitive festures, $\forall s_i \in \mathbf{s}$, $s_i^{-}$ refers to the \textit{unprivileged} group and $s_i^{+}$ to the \textit{privileged} group of the $i$-th sensitive feature.

\noindent\textbf{Proxy Features.} Given a subset of features $\mathbf{p}_i \subseteq \mathbf{x}$, if exist a function $h(\cdot)$ that can maps the relation $h:\mathbf{p}_i\mapsto s_i$, $\mathbf{p}_i$ can be considered \textit{proxy features} of the $i$-th sensitive information.

\noindent\textbf{Decision Maker and prediction.} Given an estimation function $f(\cdot)$ denoting the \textit{Decision Maker}, $\hat{y}~\in~\{0,1\}$ represents the estimation of the target label for a specific data point using only non-sensitive information, i.e. $f(\mathbf{x})=\hat{y}$.

\noindent\textbf{Sensitive-Feature Classifier and sensitive prediction.} Given an estimation function $f_{s_i}(\cdot)$ denoting the \textit{Sensitive-Feature Classifier}, $\hat{s}_i \in \{0,1\}$ represents the estimation of the $i$-th sensitive feature for a given data point, i.e. $f_{s_i}(\mathbf{x}) = \hat{s}_i$. 

\noindent\textbf{Counterfactual Generator and Counterfactuals.} Given a vector $\mathbf{x}$ and a perturbation $\mathbf{\epsilon}=\langle\epsilon_{1}, \epsilon_{2},\dots, \epsilon_{n}\rangle$, we say that a vector $ \mathbf{c}_\mathbf{x}=\langle c_{x_1}, c_{x_2},..., c_{x_n}\rangle = \mathbf{x} + \mathbf{\epsilon}$ is a counterfactual (CF) of $\mathbf{x}$ if $f(\mathbf{c}_\mathbf{x}) = 1- f(\mathbf{x})= 1- \hat{y}$. We use the set $\mathcal{C}_\mathbf{x}$, with $\lvert \mathcal{C}_\mathbf{x} \rvert = k$, to denote the set of possible \textbf{counterfactual samples} for $\mathbf{x}$. A function $g(\cdot)$ denoting the \textit{Counterfactual Generator} maps the relation between $g:\mathbf{x}\mapsto\mathcal{C}_{\mathbf{x}}$ and is used to compute $k$ counterfactuals for $\mathbf{x}$.


\noindent\textbf{Negatively-predicted samples:} Our work is focused on samples negatively predicted by the \textit{Decision Maker} (i.e., $\forall d \in \mathcal{D}$ s.t. $f(\mathbf{x})=0$) and correctly predicted by the \textit{Sensitive-Feature Classifier} (i.e.,$\forall d \in \mathcal{D}$ s.t. $f_s(\mathbf{x})=s$). For simplicity, we denote the set of such samples with $\mathcal{X}^-$, with $\mathcal{X}^- \subseteq \mathcal{D}$. For clarity, this set depends on the $f(\cdot)$ used to predict the sample and varies for each Decision Maker taken into account.

%% file: Sections/4_Methodology.tex
Our study proposes two novel metrics for detecting bias in a scenario where sensitive features are omitted (i.e., \textit{fairness under unawareness}) in the training process.
Excluding sensitive features  makes verifying that all users are treated equally incredibly challenging. 
In the instant lending case, imagine that a customer applies for a loan, and his/her request is rejected. Understanding if the customer has been discriminated is hard to verify when sensitive information is not used.
However, counterfactual reasoning can be an effective tool to propose actionable steps for reaching a positive outcome.
Our process pipeline is as follows: the \textbf{Decision Maker} makes decisions without exploiting sensitive features, then if the outcome is negative (e.g. loan rejected), the \textbf{Counterfactual Generator} is exploited to propose modifications to user characteristics and request for reaching a positive outcome (e.g. loan approved).
For each data point $d$ with a negative prediction $f(\mathbf{x}) = 0$, we generate a set of counterfactual samples $\mathcal{C}_{\mathbf{x}}$ that reach a positive outcome (i.e., $\forall \mathbf{c}_{\mathbf{x}} \in \mathcal{C}_{\mathbf{x}}$ s.t. $f(\mathbf{c}_{\mathbf{x}}) = 1$). Afterward, each counterfactual (CF) sample is evaluated by the \textbf{Sensitive-Feature Classifier} that predicts the value of the (omitted) sensitive feature for the given CF sample. 
If the CF sample is classified as e.g. male (privileged group), while the original sample was e.g. female (unprivileged group), the decision model could be biased and its unfairness can be quantified (Equations  \ref{eq:CFlipsUnpriv} and \ref{eq:DCCF}).

Indeed, each CF sample derives from the original sample $\mathbf{x}$ plus a perturbation $\epsilon$, where $\epsilon$ is the \textit{distance} from the original sample for getting a positive outcome, and it should be independent 
from the user-sensitive characteristics.
Figure \ref{fig:totalFigures} depicts a scenario in which \textit{male} (blu color) is the privileged category, and \textit{female} (red color) is the unprivileged one. 
For each subfigure, a sample with an unfavorable decision and its corresponding CFs are depicted. A classic ML model (i.e., XGB) is compared with a debiasing ML model (i.e., AdvDeb). We can observe that for the male sample and classic ML model (Figure \ref{fig:totalFigures}(a)), the CF samples belong to the same sensitive category (i.e., male). For the female sample (Figure \ref{fig:totalFigures} (b)), this is not true, revealing a bias of the model. Conversely, the debiasing model  (Figure \ref{fig:totalFigures} (c) and (d)) shows no predominance in the generated counterfactuals of one value of the sensitive class. However, a change of the outcome, e.g. from negative to positive, should not be determined by a flip of the value(s) of the sensitive feature(s).
Now, we introduce our fairness criteria and metrics.
\begin{definition}[Counterfactual Fair Opportunity]\label{def:theorem}
\textit{``A decision model is fair if the counterfactual samples of individuals with unfavorable decisions (i.e., $\mathcal{X}^-$), to reach a positive outcome (i.e., $f(\mathcal{C}_{\mathcal{X}^-})=1$), maintain the same sensitive behavior. This behavior must be guaranteed for the \textit{privileged} and for the \textit{unprivileged} group.''}
\begin{equation}\label{eq:theorem}
    \mathds{P}(f_s(\mathcal{C}_{\mathcal{X}\vert^-_{s=0}}) \neq s \mid f(\mathcal{C}_{\mathcal{X}\vert^-_{s=0}})=1, \mathcal{X}\vert^-_{s=0})=\mathds{P}(f_s(\mathcal{C}_{\mathcal{X}\vert^-_{s=1}})\neq s \mid f(\mathcal{C}_{\mathcal{X}\vert^-_{s=1}})=1, \mathcal{X}\vert^-_{s=1}) 
\end{equation}
{\normalfont \textit{which implies the complement}}:
\begin{equation}\label{eq:theorem2}
    \mathds{P}(f_s(\mathcal{C}_{\mathcal{X}\vert^-_{s=0}}) = s \mid f(\mathcal{C}_{\mathcal{X}\vert^-_{s=0}})=1, \mathcal{X}\vert^-_{s=0})=\mathds{P}(f_s(\mathcal{C}_{\mathcal{X}\vert^-_{s=1}}) = s \mid f(\mathcal{C}_{\mathcal{X}\vert^-_{s=1}})=1, \mathcal{X}\vert^-_{s=1}) 
\end{equation}
\end{definition}
Definition~\ref{def:theorem} works in a context where counterfactual samples are used for suggesting actionable steps to achieve the desired outcome. However, the outcome behaviour should not depend on a specific sensitive group. Thus, the degree of sensitive (un)fidelity of counterfactual samples must be equal between the two sensitive groups.


To define a sort of discrimination score of a given decision model, we propose a metric that we call \textit{Counterfactual Flips}. The metric quantifies the discriminatory behavior the model might put in place. 

\begin{definition}[Counterfactual Flips]
\textit{Given a sample $\mathbf{x}$ belonging to a demographic group $s$ whose model output is denoted as $f(\mathbf{x})$, generated a set $\mathcal{C}_\mathbf{x}$ of $k$ counterfactuals with desired $y^{*}$ outcome $\forall \mathbf{c}_\mathbf{x}^i \in \mathcal{C}_\mathbf{x}$ s.t. $f(\mathbf{c}_\mathbf{x}^i) = y^{*}$, the Counterfactual Flips indicate the percentage of counterfactual samples belonging to another demographic group (i.e., $f_s(\mathbf{c}_\mathbf{x}^i) \neq f_s(\mathbf{x})$, with $f_s(\mathbf{x})=s$).
\begin{equation}\label{eq:CFlips}
\mathrm{CFlips}(\mathbf{x},\mathcal{C}_{\mathbf{x}},f_s(\cdot)) \triangleq \frac{\sum_{i=1}^k(\mathbbm{1}(\mathbf{c}_\mathbf{x}^i))}{k} \quad \text{where } \mathbbm{1}(\mathbf{c}_\mathbf{x}^i) = \begin{cases}
1 &\text{if }  f_s(\mathbf{c}_\mathbf{x}^i) \neq f_s(\mathbf{x}) \neq s \\
0 &\text{if }  f_s(\mathbf{c}_\mathbf{x}^i) = f_s(\mathbf{x}) = s
\end{cases} 
\end{equation}} 
\end{definition}

The bigger the $\mathrm{CFlips}$ value is, the stronger the discriminatory bias the model suffers from.
From a probabilistic perspective, the $\mathrm{CFlips}$ can be considered as the probability of Counterfactual, generated to reach an opposite outcome from the original sample, to be predicted by the sensitive feature classifier as opposite to the original sample (see Equation~\ref{eq:CFlipsProb}).
\begin{equation}\label{eq:CFlipsProb}
    \mathds{P}(f_s(\mathcal{C}_{\mathbf{x}})\neq s \mid f(\mathcal{C}_{\mathbf{x}})=1-f(\mathbf{x}), f(\mathbf{x}), s) \quad \text{with } s \in \{0,1\}
\end{equation}
Given a set of samples $\mathcal{X}^{-} \subseteq \mathcal{D}$ predicted by the decision maker as negative (unfavorable decision, see Section~\ref{sec:preliminaries}), the metric in Equation~\ref{eq:CFlips} can be generalized to the \textit{unprivileged} and \textit{privileged} group (see Equation~\ref{eq:CFlipsUnpriv}, which is restricted to the \textit{unprivileged} samples negatively predicted, and Equation~\ref{eq:CFlipsPriv}, restricted to the \textit{privileged} samples negatively~predicted).
\begin{equation}\label{eq:CFlipsUnpriv}
    \mathrm{CFlips}_{s=0} \triangleq  \frac{\sum_{i=1}^{n}\mathrm{CFlips}(\mathbf{x}_i,\mathcal{C}_{\mathbf{x}_i},f_s(\cdot)) }{\lvert \mathcal{X}\vert_{s=0}^{-}\rvert} \quad \text{with } \mathbf{x}_i \in \mathcal{X}\vert_{s=0}^{-}
\end{equation}
\begin{equation}\label{eq:CFlipsPriv}
    \mathrm{CFlips}_{s=1} \triangleq  \frac{\sum_{i=1}^{n}\mathrm{CFlips}(\mathbf{x}_i,\mathcal{C}_{\mathbf{x}_i},f_s(\cdot)) }{\lvert \mathcal{X}\vert_{s=1}^{-}\rvert} \quad \text{with } \mathbf{x}_i \in \mathcal{X}\vert_{s=1}^{-}
\end{equation}

A limitation of the CFlips metric is that it does not measure the distance of each CF sample from the original data point. However, from an individual-fairness wise, a debated issue is the definition of a metric that considers that distance~\citep{DBLP:conf/innovations/DworkHPRZ12}.
Accordingly, we propose a new metric that considers
CFs ranked based on the Mean Absolute Deviation from the original sample and other criteria~\cite{mothilal2020dice}. The insight behind this metric is that the more the CF is ranked high (in the top positions of the ranking), the more its impact on the metric value. Thus, the metric penalizes CFs ranked in the top positions for which the value of the sensitive feature is flipped.
More formally:

\begin{definition}[Discounted Cumulative Counterfactual Fairness] Given a set of Counterfactuals $\mathbf{C}_{\mathbf{x}}$ for a sample $\mathbf{x}_i$, the \textit{Discounted Cumulative Counterfactual Fairness} $\mathrm{DCCF}_{\mathbf{x}_i}$ measure the cumulative gain of the ranking of counterfactuals with respect to the sensitive group of the original sample:
\begin{equation}\label{eq:DCCF}
    \mathrm{DCCF}_{\mathbf{x}_i} \triangleq \sum_{p_j,\mathbf{c}^j_{\mathbf{x}_i} \in \mathcal{C}_{\mathbf{x}_i}}{\frac{2^{(1 - \mathbbm{1}(c^j_{\mathbf{x}_i}))}-1}{\log_2(p_j + 1)}}
\end{equation}
where $p_j$ is the rank of $\mathbf{c}^j_{\mathbf{x}_i}$ in $\mathcal{C}_{\mathbf{x}_i}$ and $\mathbbm{1}(c^j_{\mathbf{x}_i})$ from Equation~\ref{eq:CFlips}.
\end{definition}
DCCF rewards the CF samples in the ranking that did not flip.
If more CF samples belonging to the same sensitive group as the original data point are in a higher ranking position, the result will be a higher DCCF. Thereby, we can formulate the \textit{Ideal Discounted Cumulative Counterfactual Fairness} (IDCCF) correspond to an ideal ranking in which each CF sample $\mathbf{c}_{\mathbf{x}}$ belongs to the same sensitive group as the starting sample $\mathbf{x}$ (Definition~\ref{def:IDCCF}), and the
\textit{normalized} DCCF (nDCCF) (Definition~\ref{def:nDCCF}).
\begin{definition}[Ideal Discounted Cumulative Counterfactual Fairness] The \textit{Ideal Discounted Cumulative Counterfactual Fairness} is the ideal ranking of the estimated sensitive information of counterfactuals with respect to the sensitive information of the original sample. In an ideal ranking, each counterfactual belongs to the same sensitive group of the original sample.
\end{definition}\label{def:IDCCF}
\begin{equation}\label{eq:IDCCF}
    \text{IDCCF}_{x_i} \triangleq \sum_{p_j,c^j_{x_i} \in \mathcal{C}_{x_i}}{\frac{2^{(1)}-1}{\log_2(p_j + 1)}}
\end{equation}

\begin{definition}[normalized Discounted Cumulative Counterfactual Fairness] The \textit{normalized Discounted Cumulative Counterfactual Fairness} (nDCCF) is the normalization of the current counterfactual rank DCCF with respect to the ideal rank IDCCF.
\end{definition}\label{def:nDCCF}
\begin{equation}\label{eq:nDCCF}
    \mathrm{nDCCF}_{x_i} \triangleq \frac{\mathrm{DCCF}_{x_i}}{\mathrm{IDCCF}_{x_i}}
\end{equation}






In the same way as CFlips, given a set of samples $\mathcal{X}^{-} \subseteq \mathcal{D}$ predicted by the decision maker as negative, the metric in Equation~\ref{eq:nDCCF} can be generalized to the \textit{unprivileged} and \textit{privileged} group (Equation~\ref{eq:nDCCFunpriv}~and~\ref{eq:nDCCFpriv}).
\begin{equation}\label{eq:nDCCFunpriv}
    \mathrm{nDCCF}_{s=0} \triangleq \frac{1}{|\mathcal{D}|_{s=0}|}\sum_{\mathbf{x}_i}{\mathrm{nDCCF}_{\mathbf{x}_i}} \quad \text{with } \mathbf{x}_i \in \mathcal{X}\vert_{s=0}^{-}
\end{equation}
\begin{equation}\label{eq:nDCCFpriv}
    \mathrm{nDCCF}_{s=1} \triangleq \frac{1}{|\mathcal{D}|_{s=1}|}\sum_{\mathbf{x}_i}{\mathrm{nDCCF}_{\mathbf{x}_i}} \quad \text{with } \mathbf{x}_i \in \mathcal{X}\vert_{s=1}^{-}
\end{equation}

For both $\mathrm{CFlips}$ and $\mathrm{nDCCF}$, we are interested in the difference (i.e., $\Delta\mathrm{CFlips}$ and $\Delta\mathrm{nDCCF}$), between the \textit{privileged} and \textit{unprivileged} groups, being close to zero. In fact, even though those metrics are individual-based computed 
they can serve as group-based fairness measures thanks to the $\Delta$ analysis.

%% file: Sections/5_Experiments.tex
This section provides a detailed description of our experiments.

\noindent\textbf{Dataset.} To be coherent with fairness state-of-the-art research~\cite{DBLP:conf/iclr/BalunovicRV22,DBLP:conf/nips/DoniniOBSP18,DBLP:conf/kdd/PedreschiRT08,Das33,pmlr-v28-zemel13}, we have conducted experiments on three well-known benchmark datasets (i.e., Adult~\cite{kohavi_becker_1996}, Crime~~\cite{crime_data}, and German~\cite{german_hoffman}).
\begin{enumerate}[label=(\alph*),leftmargin=*,topsep=0pt]
    \item \textit{Adult}\footnote{\url{https://archive.ics.uci.edu/ml/datasets/adult}} is a real-world popular UCI Machine Learning dataset extracted from the 1994 US Census database used for income prediction. The task is to determine if the salary of an individual is more, or less, than $50.000\$$. In the dataset, we have different sensitive information. In our work, we consider only \textit{gender}, indicating the sex of the individual (i.e., \textit{male} or \textit{female}), and \textit{marital-status}, indicating if is married or not.
    \item \textit{Adult-Debiased} is our \textit{pre-processed} version of the \textit{Adult} dataset. Since the Adult dataset presents other sensitive features (e.g., \textit{age}, \textit{relationships}, and \textit{race}), we removed all these features. Furthermore, we remove all the features that are highly correlated with at least one sensitive information (i.e., with a Pearson's correlation equal or greater than 0.35). In addition, we condensed the category of the feature \textit{work-class} into three general classes to avoid specialization of sensitive information with a specific category (i.e., we replaced the categories \textit{Private}, \textit{SelfEmpNotInc}, \textit{SelfEmpInc}, with \textit{Private}, the categories \textit{FederalGov}, \textit{LocalGov}, \textit{StateGov}, with \textit{Public}, and the category \textit{WithoutPay} with \textit{Unemployed}). The task and sensitive information considered are the same as for the \textit{Adult} one.
    \item \textit{Crime}\footnote{\url{https://archive.ics.uci.edu/ml/datasets/US+Census+Data+(1990)}} is a real-world UCI Machine Learning Census dataset for violent state prediction. The task is to predict if a state can be considered violent due to the number of crimes greater than the median of all states' crimes. We reproduce the setting of \citet{DBLP:conf/iclr/BalunovicRV22} and we consider as sensitive information the \textit{race} representing the ethnicity with the most number of crimes in that particular state and is divided into two groups, i.e., \textit{white}, as the privileged, and \textit{other}, as the unprivileged group.
    \item \textit{German}\footnote{\url{https://archive.ics.uci.edu/ml/datasets/statlog+(german+credit+data)}} is a real-world UCI Machine Learning dataset extracted from a German bank for default prediction. The task is to predict if an individual, based on his characteristics and on the credit he requires, will or not default on the loan.
    As for the \textit{Adult} dataset, we have several sensitive information, we decided to focus only on \textit{gender}, indicating the sex of the individual (i.e., \textit{male} or \textit{female}), and \textit{age}, if an individual is older, or younger, than 25 years.
\end{enumerate}
As a general reminder, the above-mentioned sensitive information  are not included in the training of the dataset to be coherent with the \textit{fairness under unwareness} setting. Additional information on the
target and sensitive-feature distribution of the datasets are available in Tabel~\ref{tab:datasetSplit}.
\input{tables/DatasetStatistics}

\noindent\textbf{Decision Maker.}
To keep the approach as general as possible, 
we opted for different Machine learning models\footnote{\underline{LR}, \underline{SVM}, \underline{DT}, \underline{RF}, \underline{MLP}: \url{https://scikit-learn.org/}; \underline{XGB}: \url{https://github.com/dmlc/xgboost}; \underline{LGBM}: \url{https://github.com/microsoft/LightGBM}}: linear model, i.e., Logistic Regression (LR), decision boundary-based model, i.e., Support-Vector Machines (SVM), tree-based models, i.e., Decision Tree (DT), Random Forest (RF),  XGBoost (XGB)
, and LightGBM (LGBM), and neural model, i.e., Multi-Layer Perceptron (MLP). 

\noindent\textbf{Debiased Decision Maker.} To investigate the quality and the reliability of our metrics we used also two debiased classifier\footnote{\underline{AdvDeb}: \url{https://github.com/Trusted-AI/AIF360}; \underline{LFERM}: \url{https://github.com/jmikko/fair_ERM}; \underline{FairC} : \url{https://github.com/mbilalzafar/fair-classification}}, \textit{Adversarial Debiasing} (AdvDeb) proposed by \citet{10.1145/3278721.3278779}, \textit{Linear Fair Empirical Risk Minimization} (LFERM) proposed by \citet{DBLP:conf/nips/DoniniOBSP18}, and \textit{Fair Classification} (FairC) proposed by \citet{pmlr-v54-zafar17a}. The choice is not arbitrary as they are in-processing algorithms that use sensitive information only in the training phase and can handle unawareness in the inference phase.
As a further point to their advantage, they are black box models that can be interpreted in contrast to models belonging to \textit{Learning Fair Representation} domain which maps data into a space that can no longer be interpreted.

\noindent\textbf{Counterfactual Generator.}
For the sake of reproducibility and reliability, the counterfactuals are generated with an external counterfactual framework, DiCE~\cite{mothilal2020dice,10.1145/3351095.3372850}. DiCE is an open-source framework developed by Microsoft that offers several strategies for generating diverse and plausible candidate counterfactual samples. The framework is model-agnostic and manages the generation of diverse counterfactual samples that satisfy \textit{Proximity}, \textit{Sparsity}, \textit{Diversity}, and \textit{Feasibility} constraints.
We choose to use two different strategies: the Genetic strategies which generate samples that belong to the unexplored space to investigate the model behavior with new samples (i.e., $\mathbf{c}_\mathbf{x} \notin \mathcal{D}$), and KDtree which generate samples that belong to the known space (i.e., $\mathbf{c}_\mathbf{x} \in \mathcal{D}$). We use also another model-agnostic counterfactual generator (i.e., MACE from \citet{DBLP:conf/aistats/KarimiBBV20} which is based on first-order logic). However, the framework is compatible with only LR, DT, RF, and MLP (only for a 10-neuron single layer) and never handles the generation of required diverse counterfactual samples for each dataset making the comparison unfeasible. Indeed, the number of counterfactuals queried for analysis purposes is equal to 100 (i.e., $|\mathcal{C}_\mathbf{x}|=k=100$). More statistics about counterfactual tool generation are available in Appendix~\ref{sec:appendix}.


\noindent\textbf{Sensitive-Feature Classifier.}
This component is essential to our methodology since it enables the algorithm to find discriminatory models through our metrics. Since the performance of this step can compromise the reliability of our metrics, we investigate different \textit{sensitive-feature classifiers}: RF, MLP, and XGB.


\noindent\textbf{Metrics.} We evaluate the models performance with the Area~Under~the~Receiver~Operative~Curve~(AUC), Accuracy (ACC), Recall, Precision, F1 score, and fairness measures of Statistical Parity\footnote{$DSP = \abs{\mathds{P}(\hat{Y}=1 \mid S=1) - \mathds{P}(\hat{Y}=1 \mid S=0)}$} (DSP), Equal Opportunity\footnote{$DEO = \abs{\mathds{P}(\hat{Y}=1 \mid S=1,Y=1) - \mathds{P}(\hat{Y}=1 \mid S=0,Y=1)}$} (DEO), and Average Odds\footnote{$DAO = \frac{1}{2}\abs{\sum_{Y \in \{0,1\}}(\mathds{P}(\hat{Y}=1 \mid S=1,Y) - \mathds{P}(\hat{Y}=1 \mid S=0,Y))}$} (DAO) as a preliminary fairness analysis.

\noindent\textbf{Split and Hyperparameter Tuning.} The datasets have been split with the hold-out method 90/10 train-test set, with stratified sampling w.r.t. the target and sensitive labels, to respect original distribution in each split. 
The ``Decision Maker'', the ``Debiased Decision Maker'', and the ``Sensitive-Feature Classifier'' models have been tuned on the training set with a Grid Search k-fold (k=5) cross-validation methodology, the first two optimizing AUC metric, and the latter F1 score to prevent unbalanced predictions on the sensitive feature.

%% file: tables/DatasetStatistics.tex
\begin{table}[]
    \centering
    \caption{Overview of relevant dataset information, including sensitive feature distribution, target distribution, name of privileged group, and ex-ante Statistical Parity.}\label{tab:SFdataDisrib}\label{tab:datasetSplit}
\begin{subtable}{\textwidth}
\begin{tabular}{cccccccccc}\hline\toprule
Dataset&$\lvert \mathcal{D} \rvert$&$\lvert n \rvert$  & $Y$ &$Y=1$& $s$ & $s=1$& $\Phi(s)^\dagger$ &$\Phi(Y)^{\dagger\dagger}$&ex-ante SP$^*$ \\
\midrule
Adult & 45222&13 &income&$\geq \$50k$&gender &male &0,675/0,325 & 0,248/0,752 &0.199 \\
Adult-deb. &45222 &6 &income&$\geq \$50k$ &gender &male &0,675/0,325 & 0,248/0,752 &0.199 \\
Crime & 1994&98 &Violent State&$<$\textit{median}&race &white &0,58/0,42 & 0,5/0,5 &0.554 \\
German&1000 &17 &credit score&Good&gender &male &0,690/0,310 & 0,7/0,3 &0.075 \\
\bottomrule 
\hline
\end{tabular}
\end{subtable}


\begin{subtable}{\textwidth}
\begin{tabular}{ccccc}

\multicolumn{5}{l}{\footnotesize $^\dagger$ Probability distribution of the \textit{privileged} and \textit{unprivileged} group: $\mathds{P}(S=1)/\mathds{P}(S=0)$} \\
\multicolumn{5}{l}{\footnotesize $^{\dagger\dagger}$ Probability distribution of the target variable: $\mathds{P}(Y=1)/\mathds{P}(Y=0)$} \\
\multicolumn{5}{l}{\footnotesize $^*$ A priori Statistical Parity probability, based on Independence Statistical Criteria: $\mathds{P}(Y=1 \mid S=1) - \mathds{P}(Y=1 \mid S=0)$} 
\end{tabular}
\end{subtable}
\vspace{-4mm}
\end{table}

%% file: Sections/6_Results.tex
This Section introduces and discusses the experimental results to assess the effectiveness of the proposed model and metrics. 
Due to the lack of space, we focus the analysis of the fairness metrics (Table~\ref{tab:totResCFLIPNDCCF} and Table~\ref{tab:totKDtreeCFLIPNDCCF}) on one $f_s(\cdot)$, i.e., XGB, and one sensitive information, i.e., \textit{gender} for Adult, Adult-debiased, and German datasets, and \textit{race} for Crime dataset. Nevertheless, in Section~\ref{sec:subSec3}, we conduct a comprehensive comparison including all $f_s(\cdot)$. The extended results are available in Appendix~\ref{sec:appendix}. 

\subsection{Decision Makers and Sensitive-feature classifiers performance comparison}\label{sec:subSec1}
\input{tables/SFxgbTable.tex}
The first experiment (see Table~\ref{tab:SFandCLF}) is a two-step evaluation to establish whether the sensitive-feature classifier and the decision maker provide reliable outcomes.
The first evaluation step assesses how well the sensitive-feature classifier can determine whether an individual belongs to the privileged or unprivileged group. For each dataset, we train a sensitive-feature classifier for gender or race according to the dataset.
The second evaluation step offers a broad overview of the performance of the Decision Makers. Let us report the most relevant observations:

\begin{itemize}[leftmargin=*,topsep=0pt]
\item  
The first remark is that the \textit{Sensitive-Feature Classifiers} perform well across all datasets. This result suggests that the classifiers can infer sensitive information from other features.
Another important observation is that sensitive-feature classifiers exhibit worse performance on the Adult-debiased than on the original Adult dataset. 
This behavior is probably due to the debiasing process, as the Adult-debiased dataset does not contain the features highly correlated with the sensitive ones. 
Interestingly, no model consistently outperforms the others. 
Let us take AUC as an example. 
There is not a model that overall outperforms the others.
However, in a scenario characterized by imbalanced classes, a model should exhibit high AUC and F1 to provide accurate predictions and separate \textit{privileged} from \textit{unprivileged} samples. 
\item Regarding the Decision Makers, Table~\ref{tab:SFandCLF} highlights that tree-based models, especially XGB, generally achieve the best performance. Conversely, the Debiasing models (i.e., AdvDeb, LFERM, and FairC) perform worse than standard models. 
However, as expected, adopting \textit{Fairness Under Unawareness} -- i.e., removing all the sensitive information and, for Adult-debiased, also removing highly correlated features -- has caused a worsening of the performance for all the Decision Makers (see the comparison between Adult and Adult-debiased results). 
This observation suggests that sensitive and sensitive-correlated information can improve the prediction of the target label, indicating bias.
\end{itemize}
\noindent \underline{Final comments.}
\textit{Experiments show that sensitive-feature classifiers can predict the label even when features poorly correlated with sensitive information are available. That indicates that the Fairness Under Unawareness setting is insufficient to protect individuals from discrimination.
In addition, Decision Makers' performance decreases when sensitive information are excluded. Since they learn from real-world datasets that most likely embed
human bias, this suggests that sensitive features influence the original human decision-making process.}


\subsection{Fairness, CFlips, and nDCCF results}\label{sec:subSec2}
\input{tables/CFlipsGeneticNeg.tex}
\input{tables/CFlipsKDtreeNeg.tex}
Now that we have a clear vision of each classifier's performance, we can move on to analyze how well they perform in terms of fairness. 
The performance of the Decision Makers on the metrics of DSP, DEO, and DAO, as well as our suggested metrics $\mathrm{CFlips}$ and $\mathrm{nDCCF}$ are reported in Table~\ref{tab:totResCFLIPNDCCF}. 
It is important to point out that the CFlips metric indicates how often a change of result for the Decision Maker corresponds to a change in the classification of the sensitive feature (e.g., from female to male and vice-versa). Conversely, the nDCCF metric gives more importance to counterfactuals with highest positions in the ranking (the most similar to the original sample) that do not change the sensitive class.
\begin{itemize}[leftmargin=*,topsep=0pt]
    \item Considering the Difference in Statistical Parity (DSP),  Difference in Equal Opportunity (DEO), and Difference in Average Odds (DAO) in Table~\ref{tab:totResCFLIPNDCCF}, it is noteworthy to mention that eliminating race and gender as sensitive information has not reduced model discrimination.  
    This result shows that proxy features makes the \textit{Fairness Under Unawareness} setting useless since the model can implicitly learn sensitive information from them. 
    The Adult-debiased dataset, where DSP and DAO results are typically better than on the Adult dataset, is a clear example.
    Indeed, due to non-linear proxy features, discrimination still exists to some extent.
    Furthermore, the debiased \textit{Decision Makers} (i.e., AdvDeb, lferm, FairC) do not appear to consistently enhance fairness performance despite implementing the in-processing debiasing constrained optimization. 
    \item Considering the $\mathrm{CFlips}$ metric we can notice that  for the \textit{unprivileged} group we generally have higher values. This indicates that samples belonging to the unprivileged group need to take the characteristics of the privileged group to reach a positive outcome (i.e., $f(\mathbf{c}_\mathbf{x})=1$). This is confirmed by the $\mathrm{nDCCF}$ metric where
    we can see how the (most similar) counterfactuals of the unprivileged  group should take on the characteristics of the privileged group to pass to a favorable prediction.
    \item From a group-based point of view, we are interested in evaluating the difference of the two proposed metrics between the privileged and unprivileged group (i.e., $\Delta\mathrm{CFlips}$ and $\Delta\mathrm{nDCCF}$) and see for which model the value of the $\Delta$ is near to zero. 
    We point out that for models with highly accurate sensitive feature classifiers (i.e., Adult and Crime), debiasing models seem to perform best in Fairness and thus succeed in fulfilling their debiasing task. This is confirmed by both standard fairness metrics and our new proposed metrics ($\Delta\mathrm{CFlips}$ and $\Delta\mathrm{nDCCF}$).  In contrast, debiasing models rarely give the best results when considering datasets with low correlation with sensitive information (i.e., Adult-debiased and German - with worst sensitive classifiers perfomance). As a matter of fact, for Adult-debiased, SVM performs better than other debiasing models. Indeed, $\Delta\mathrm{CFlips}$ and $\Delta\mathrm{nDCCF}$, differently from standard fairness metrics, reward SVM as the best-performing model.
    A similar trend can be seen in German.
    However, since there are no correctly predicted female samples for MLP and LFERM models, we have 0 $\mathrm{CFlips}$ for the \textit{unprivileged} group. Thus, the small size of the dataset made the evaluation of the metrics impracticable.
    \item Table~\ref{tab:totKDtreeCFLIPNDCCF} reports results of our metrics with counterfactuals generated with the KDtree strategy. 
   KDtree generates counterfactuals by choosing from samples that already belong to the dataset. Therefore, unlike Genetic, which generates new samples, KDtree does not explore an unknown space.
    This means that each counterfactual is chosen from the known data space (i.e., $\mathbf{c}_\mathbf{x} \in \mathcal{D}$), so KDtree measures how similar samples behave. We can see that the trend is similar to the Genetic strategy but with higher $\mathrm{CFlips}$ and $\mathrm{nDCCF}$. In this case, enlarging the number of counterfactuals for each sample worsens the metric values 
    since more distant counterfactual samples are choosen.
\end{itemize}
\noindent\underline{Final comments.} \textit{The classifiers seem to be affected by discrimination even when the sensitive information is omitted. 
Accordingly, this further confirms that the \textit{Fairness Under Unawareness} setting is insufficient to avoid decision biases and discrimination. It becomes clear that unprivileged samples must adopt privileged samples' traits to get favorable decisions.
Furthermore, this experiment shows that counterfactual reasoning is an efficient tool for identifying discriminatory biases.}

\subsection{Detecting proxy features through counterfactuals reasoning}\label{sec:subSec4}
\begin{table}[!t]\centering
\caption{Demonstrative example of $\rho$ computation based on $\mathcal{E}$ and $\Delta$ for a numerical, ordinal, and categorical feature of the Adult-debiased dataset.}\label{tab:Correlation}
\renewcommand{\arraystretch}{0.5}
\begin{tabular}{lrrrrr|lr}\hline\toprule
&numeric &ordinal&\multicolumn{3}{c}{Category (\textit{workclass})} & & \textit{gender} \\\cmidrule(l){2-2}\cmidrule(l){3-3}\cmidrule(lr){4-6}\cmidrule(l){8-8}
&\textit{capital gain} & \textit{education-num}&\textit{Private} &\textit{Public} &\multicolumn{1}{c}{\textit{Unemployed}} & &$\mathds{E}[f_s(\mathbf{x})|\mathbf{x}]$ \\
\midrule
$\mathbf{c}_{\mathbf{x}_1}$ &5000 & 6&1 &0 &0 &$f_s(\mathbf{c}_{\mathbf{x}_1})$&0.7 \\
$\mathbf{x}_1$ &2000 & 2&0 &0 &1 &$f_s(\mathbf{x}_1)$ &0.1 \\ \midrule
$\mathbf{\epsilon_{\mathbf{x}_1}}$ &3000 &4 &1 &0 &-1 &$\delta_{\mathbf{x}_1}$&0.6 \\ \midrule $\mathbf{c}_{\mathbf{x}_2}$ &2800 & 5&0 &1 &0 &$f_s(\mathbf{c}_{\mathbf{x}_2})$&0.3 \\
$\mathbf{x}_2$ &600 &5&1 &0 &0 &$f_s(\mathbf{x}_2)$ &0.7 \\ \midrule
$\mathbf{\epsilon_{\mathbf{x}_2}}$ &2200 & 0 &-1 &1 &0 &$\delta_{\mathbf{x}_2}$&-0.4 \\ \midrule
\multicolumn{7}{c}{$\dots$}\\ \midrule
$\mathbf{\epsilon_{\mathbf{x}_3}}$ &1200 &-1&0 &1 &-1 &$\delta_{\mathbf{x}_3}$&-0.6 \\ \midrule
$\rho(\mathcal{E},\Delta)$ & 0.91&0.93 &0.78& -0.99& \multicolumn{1}{r}{-0.36} &  \\ 
\bottomrule
\hline
\end{tabular}
\end{table}

\input{groupplot/proxyfeature.tex}

As a further contribution, we want to identify the features that drive the Decision Maker to a positive outcome and that, at the same time, lead to a Flip in the sensitive information. We study this relationship by investigating the Pearson correlation between the $\epsilon$ (i.e., $\mathbf{c}_\mathbf{x}-\mathbf{x}$) and the distance $\delta$ of the model expected prediction of the counterfactual and of the sample (i.e., $\delta = \mathds{E}[f_s(\mathbf{c}_{\mathbf{x}})|\mathbf{c}_\mathbf{x}] - \mathds{E}[f_s(\mathbf{x})|\mathbf{x}]$). For a numerical or ordinal feature $i$, $\epsilon_i$ can be expressed as the difference between the counterfactual and the feature of the sample  $c_{x_i} - x_i$. For a categorical feature $j$, $\epsilon_j$ can be expressed in a \textit{one-hot encoding} form as -1 to the category that is removed and 1 to the category that is engaged. We identify with $\mathcal{E}$ and $\Delta$, respectively, the set of all perturbations of $\mathcal{C}_{\mathbf{x}}$ and the difference between all the expected values of $\mathds{E}[f_s(\mathcal{C}_{\mathbf{x}})|\mathcal{C}_{\mathbf{x}}]$ and $\mathds{E}[f_s(\mathbf{x})|\mathbf{x}]$, in both cases
$\forall \mathbf{x} \in \mathcal{X}^-$.
Thus, it is possible to identify the most influential features for $f_s(\cdot)$ evaluating the Pearson correlation between $\mathcal{E}$ and $\Delta$: $\rho(\mathcal{E}, \Delta)$ (a demonstrative example in Table~\ref{tab:Correlation}).
\begin{itemize}[leftmargin=*,topsep=0pt]
\item In Figure~\ref{fig:explainCorr} we can find the rank of features correlation  with a Flip in $f_s(\cdot)$ with MLP as $f(\cdot)$ decision boundary for the generation of $\mathbf{c}_\mathbf{x}$ and XGB as $f_s(\cdot)$ for the Adult-debiased dataset. The analysis is restricted to only samples negatively predicted in order to specifically quantify the \textit{proxy-features} that lead to a positive prediction with also a change in the sensitive information. In detail, a negatively correlated feature (e.g., \textit{Adm-Clerical}) is a feature that has an opposite direction with respect to $\mathds{E}[f_s(\mathcal{X}^-) \mid \mathcal{X}^-]$ while a positively correlated one (e.g., \textit{Craft-repair}) has the same direction. 
\end{itemize}
\noindent\underline{Final comments.} \textit{Counterfactual reasoning not only can accurately detect and quantify biases in the decision process but also can quantify the contribution of each feature with a positive, or negative, outcome. }

\subsection{Ablation on the number of counterfactuals and the sensitive-feature classifier}\label{sec:subSec3}
\input{groupplot/ablationPlots.tex}
In Section~\ref{sec:subSec2}, we have evaluated the performance of the models based on our proposed metrics. However, these metrics have been evaluated only with $k$ equal to 10, 50, and 100 and only with XGB as the sensitive-feature classifier. Now, we study the effect of the number of generated counterfactuals on different sensitive-feature classifiers (i.e., RF and MLP). Furthermore, nDCCF can be affected by the ranking criterion of the counterfactual generator. This point deserves a broader discussion as follows.

As mentioned in Section \ref{sec:preliminaries}, a counterfactual sample $\mathbf{c}_\mathbf{x}$ is a deviation from a starting vector $\mathbf{x}$ of a quantity $\mathbf{\epsilon}$ that is computed $k$ times (the number of counterfactuals).
For each sample, using a function $g(\cdot)$, we generate a set of counterfactuals $\mathcal{C}_\mathbf{x}$ such that $g(\mathbf{x})=\mathcal{C}_\mathbf{x}$. 
This set is ranked according to a model-specific utility function $u(\cdot)$ (e.g., euclidean distance or absolute distance of the counterfactual sample from the original one\footnote{A counterfactual that is closer to the original sample has greater utility than one further away.}). 
Indeed, $g(\mathbf{x})$ returns a set of counterfactuals such that $\mathcal{C}_\mathbf{x}=\langle \mathbf{c}^1_\mathbf{x}, \mathbf{c}^2_\mathbf{x}, \dots, \mathbf{c}^k_\mathbf{x} \rangle$ with $u(\mathbf{c}^i_\mathbf{x})>u(\mathbf{c}^{j}_\mathbf{x})$ and $i<j$.
However, the ranking of this set depends on the strategy used by the \textit{counterfactual generator}. 
In this regard and to be totally agnostic from that strategy, we reranked the list of counterfactual samples based on the absolute difference between the expected model prediction of a counterfactual sample and that of the original sample (i.e., $u(\mathbf{c}_\mathbf{x}) = -|\mathds{E}[f(\mathbf{x})|\mathbf{x}] - \mathds{E}[f(\mathbf{c}_\mathbf{x})|\mathbf{c}_\mathbf{x}]|$, s.t. $\forall \mathbf{c}^{i}_\mathbf{x},\mathbf{c}^{j}_\mathbf{x} \in \mathcal{C}_\mathbf{x}$, with $i<j$, $u(\mathbf{c}^i_\mathbf{x})>u(\mathbf{c}^{j}_\mathbf{x})$). Due to space constraints, we narrow our analysis to the Adult dataset (see Figure \ref{fig:ablationAdult}).
\begin{itemize}[leftmargin=*,topsep=0pt]
    \item Comparing the different sensitive-feature classifiers, it is evident that the metrics can be considered stable also due to the high performance of each classifier. Furthermore, considering the $\Delta\mathrm{CFlips}$ and $\Delta\mathrm{nDCCF}$, we can notice that for each Decision Maker, we reach a stable result after 20 generated counterfactuals except for LFERM. LFERM seems to increase the value of each metric by enlarging the number of counterfactual samples. Investigating motivations from a distance perspective may be a viable option, but it is a current challenge and limitation of fairness research~\cite{DBLP:conf/innovations/DworkHPRZ12}. Another interesting point is the similarity between the two metrics. It seems that the trend of the $\Delta\mathrm{CFlips}$ with the increasing of $|\mathcal{C}_\mathbf{x}|$ is consistent with the one of $\Delta\mathrm{nDCCF}$.
    \item Considering the sorted version of $\Delta\mathrm{nDCCF}$, we can observe three different trends: (i) the Decision Maker has a behavior similar  to the sorted version meaning that the discrimination is also in the proximity of the positive decision boundary side, (ii) models like LFERM and FairC, instead, have an opposite behavior in the proximity of the positive decision boundary, and (iii) AdvDeb starts with greater discrimination  for counterfactual samples closer to the decision boundary and then becomes fairer with the distance increasing.
\end{itemize}
\noindent\underline{Final comments.} \textit{
The findings show that counterfactual reasoning is an appropriate tool for identifying decision-making biases and for enhancing SOTA fairness indicators.
Studying classifiers' decision boundaries can also reveal further information about the discrimination behavior of the model. 
}

%% file: tables/SFxgbTable.tex
\begin{table}[!t]\centering
\caption{AUC, Accuracy, Precision, Recall, and F1 score on the Adult, Adult-debiased,
Crime, and German test set of the Sensitive Feature Classifiers and Target Classifiers. We
mark the best-performing model for each metric in bold font.}\label{tab:SFandCLF}
\scriptsize
\rowcolors{5}{gray!15}{white}
\begin{adjustbox}{width=\textwidth, center}
\begin{tabular}{lllrrr|rrrrrrrrrr}
\hline
\toprule
& & &\multicolumn{3}{c}{\textbf{Sensitive Feature Classifier}} &\multicolumn{10}{c}{\textbf{Target Classifier}} \\
\cmidrule(lr){4-6} \cmidrule(lr){7-16}
Dataset &s &metric $\uparrow$ &\textbf{RF} &\textbf{MLP} &\multicolumn{1}{c}{\textbf{XGB}} &\textbf{LR} &\textbf{DT} &\textbf{SVM} &\textbf{LGB} &\textbf{XGB} &\textbf{RF} &\textbf{MLP} &\textbf{LFERM} &\textbf{AdvDeb} &\textbf{FairC} \\\midrule
\cellcolor{white}&\cellcolor{white} &AUC &0.9402 &0.9363 &\textbf{0.9413} &0.9078 &0.8484 &0.9073 &0.9304 &\textbf{0.9314} &0.9118 &0.9119 &0.9031 &0.9123 &0.8770 \\
& &ACC &0.8539 &\textbf{0.8559} &0.8463 &0.8099 &0.8161 &0.8541 &0.8658 &\textbf{0.8698} &0.8534 &0.8494 &0.8428 &0.8512 &0.8395 \\
\cellcolor{white}&\cellcolor{white} &Precision &0.9043 &0.9065 &\textbf{0.9549} &0.5782 &0.6879 &0.7570 &0.7655 &\textbf{0.7737} &0.7371 &0.7222 &0.7324 &0.7500 &0.7382 \\
& &Recall &0.8762 &\textbf{0.8768} &0.8107 &\textbf{0.8608} &0.4719 &0.6057 &0.6610 &0.6708 &0.6351 &0.6378 &0.5763 &0.5995 &0.5459 \\
\multirow{-5}{*}{\cellcolor{white}Adult} &\multirow{-5}{*}{\cellcolor{white}gender} &F1 &\textbf{0.8900} &0.8914 &0.8769 &0.6918 &0.5598 &0.6729 &0.7094 &\textbf{0.7186} &0.6823 &0.6774 &0.6450 &0.6663 &0.6277 \\ \midrule
\cellcolor{white}&\cellcolor{white} &AUC &\textbf{0.8028} &0.8010 &0.7896 &0.8233 &0.7895 &0.7944 &\textbf{0.8596} &0.8578 &0.8336 &0.8271 &0.8017 &0.8309 &0.7981 \\
\cellcolor{white}& \cellcolor{white}&ACC &\textbf{0.7482} &0.7480 &0.7444 &0.7367 &0.8017 &0.8061 &0.8371 &\textbf{0.8375} &0.8267 &0.8156 &0.7953 &0.8196 &0.8054 \\
& &Precision &0.7699 &0.7832 &\textbf{0.8111} &0.4790 &0.8294 &\textbf{0.8389} &0.8038 &0.8063 &0.7621 &0.7540 &0.7079 &0.7529 &0.7526 \\
\cellcolor{white}&\cellcolor{white} &Recall &\textbf{0.8942} &0.8664 &0.8100 &\textbf{0.7119} &0.2516 &0.2694 &0.4532 &0.4532 &0.4371 &0.3800 &0.2962 &0.4050 &0.3202 \\
\multirow{-5}{*}{\cellcolor{white}Adult-deb} &\multirow{-5}{*}{\cellcolor{white}gender} &F1 &\textbf{0.8274} &0.8227 &0.8106 &0.5727 &0.3860 &0.4078 &0.5796 &\textbf{0.5802} &0.5556 &0.5053 &0.4176 &0.5267 &0.4493 \\ \midrule
\cellcolor{white}&\cellcolor{white} &AUC &0.9893 &0.9885 &\textbf{0.9910} &0.9248 &0.8991 &\textbf{0.9288} &0.9168 &0.9099 &0.9096 &0.9203 &0.9100 &0.9008 &0.8024 \\
& &ACC &0.9450 &\textbf{0.9500} &0.9450 &\textbf{0.8700} &0.8200 &\textbf{0.8700} &0.8400 &0.8500 &0.8400 &0.8650 &0.8400 &0.8100 &0.7500 \\
\cellcolor{white}&\cellcolor{white} &Precision &0.9412 &\textbf{0.9417} &0.9412 &0.8627 &0.8265 &\textbf{0.8776} &0.8400 &0.8500 &0.8400 &0.8544 &0.8333 &0.8444 &0.7500 \\
& &Recall &0.9655 &\textbf{0.9741} &0.9655 &\textbf{0.8800} &0.8100 &0.8600 &0.8400 &0.8500 &0.8400 &\textbf{0.8800} &0.8500 &0.7600 &0.7500 \\
\multirow{-5}{*}{\cellcolor{white}Crime} &\multirow{-5}{*}{\cellcolor{white}race} &F1 &0.9532 &\textbf{0.9576} &0.9532 &\textbf{0.8713} &0.8182 &0.8687 &0.8400 &0.8500 &0.8400 &0.8670 &0.8416 &0.8000 &0.7500 \\ \midrule
& &AUC &0.7106 &0.5091 &\textbf{0.7139} &\textbf{0.8186} &0.7219 &0.8110 &0.7614 &0.7871 &0.7936 &0.8162 &0.7605 &0.7371 &0.8152 \\
\cellcolor{white}&\cellcolor{white} &ACC &\textbf{0.7300} &0.6900 &0.6900 &0.7600 &0.7600 &0.7600 &0.7500 &\textbf{0.7900} &0.7600 &0.7600 &0.7200 &0.7300 &0.7400 \\
& &Precision &0.7234 &0.6900 &\textbf{0.7879} &\textbf{0.8485} &0.7805 &0.7738 &0.7848 &0.8025 &0.7674 &0.7738 &0.7188 &0.7792 &0.7619 \\
\cellcolor{white}&\cellcolor{white} &Recall &0.9855 &\textbf{1.0000} &0.7536 &0.8000 &0.9143 &0.9286 &0.8857 &0.9286 &0.9429 &0.9286 &\textbf{0.9857} &0.8571 &0.9143 \\
\multirow{-5}{*}{German} &\multirow{-5}{*}{gender} &F1 &\textbf{0.8344} &0.8166 &0.7704 &0.8235 &0.8421 &0.8442 &0.8322 &\textbf{0.8609} &0.8462 &0.8442 &0.8313 &0.8163 &0.8312 \\

\bottomrule
\hline
\end{tabular}
\end{adjustbox}
\end{table}

%% file: tables/CFlipsGeneticNeg.tex
\begin{table*}[t]
    \centering
    \scriptsize
    \caption{(GENETIC) DSP, DEO, and DAO results on Test set; \text{CFlip} and \text{nDCCF} results at different $|k|$ number of Counterfactuals for each negatively predicted Test set sample ($0^*$ there are no negative predicted \textit{unprivileged} samples which result in no CF samples for the unprivileged group). We mark the best-performing model for each fairness metric in bold font.}
    \label{tab:totResCFLIPNDCCF}
    \begin{adjustbox}{width=\textwidth, center}
\setlength{\tabcolsep}{2.5pt}
\renewcommand{\arraystretch}{1}
\rowcolors{12}{gray!15}{white}
\begin{tabular}{lllrrr|rrr|rrr|rrr|rrr|rrr|rrr}
\hline
\toprule
& & &  & & \multicolumn{1}{c}{}  & \multicolumn{9}{c}{$\text{CFlips}$@$|k|$ (\%)} & \multicolumn{9}{c}{$\text{nDCCF}@|k|$}
\\ \cmidrule(lr){7-15} \cmidrule(lr){16-24}
& & &  & &  \multicolumn{1}{c}{} & \multicolumn{3}{c}{Privileged} & \multicolumn{3}{c}{Unprivileged} & \multicolumn{3}{c}{$\Delta\text{CFlips}\downarrow
$}&\multicolumn{3}{c}{Privileged} & \multicolumn{3}{c}{Unprivileged} & \multicolumn{3}{c}{$\Delta\text{nDCCF}\downarrow
$} \\ \cmidrule(lr){7-9} \cmidrule(lr){10-12} \cmidrule(lr){13-15} \cmidrule(lr){16-18} \cmidrule(lr){19-21} \cmidrule(lr){22-24}  
 Dataset   &$s$ & model  &   DSP$\downarrow$ & DEO$\downarrow$
 &   \multicolumn{1}{c}{DAO$\downarrow$
} &   @10 &   @50 &   \multicolumn{1}{c}{@100} &   @10 &   @50 &   \multicolumn{1}{c}{@100} &   @10 &   @50 &   \multicolumn{1}{c}{@100} &   @10 & @50 & \multicolumn{1}{c}{@100} & @10 &   @50 &   \multicolumn{1}{c}{@100} & @10 & @50 &   @100 \\
\midrule
     && \textbf{LR}       &     0.2947 &     0.0546 & 0.1241  &     12.332 &        10.886 &         10.212 &        66.353 &        72.932 &         77.165 &        54.021 &        62.046 &         66.953 &        0.8678 &        0.8849 &         0.886  &        0.3522 &        0.2913 &         0.2497 &               0.5156 &               0.5936 &                0.6363 \\
 \cellcolor{white}  & \cellcolor{white}  & \textbf{DT}       &0.1461 & 0.0760& 0.0722&     8.721 &          9.442 &           9.563 &         67.553 &         73.179 &          74.152 &        58.832 &        63.737 &         64.589 &        0.911  &        0.9067 &         0.8988 &        0.3371 &        0.284  &         0.2685 &               0.5739 &               0.6227 &                0.6303 \\
  &  & \textbf{SVM}       &     0.1769 &     0.0644 & 0.0692   &     6.752 &         7.533 &          7.742 &        77.095 &        80.973 &         81.372 &        70.343 &        73.44  &         73.63  &        0.9306 &        0.9258 &         0.9171 &        0.2474 &        0.2042 &         0.1948 &               0.6832 &               0.7216 &                0.7223 \\
\cellcolor{white}  &\cellcolor{white}   & \textbf{LGBM}       &     0.1850 &     0.0379 &  0.0569  &     9.195 &         8.541 &          8.781 &        65.918 &        76.605 &         79.697 &        56.723 &        68.064 &         70.916 &        0.9049 &        0.9124 &         0.9049 &        0.3611 &        0.2633 &         0.2272 &               0.5438 &               0.6491 &                0.6777 \\
 &  & \textbf{XGB}       &     0.1884 &     0.0635 &   0.0680  &   10.011 &         8.788 &          9.07  &        64.796 &        76.243 &         79.512 &        54.785 &        67.455 &         70.442 &        0.8968 &        0.9088 &         0.9014 &        0.3708 &        0.2677 &         0.2298 &               0.526  &               0.6411 &                0.6716 \\
\cellcolor{white}  &\cellcolor{white}  & \textbf{RF}       &0.1854&0.0216&0.0545&     7.18  &          7.226 &           7.577 &         68.926 &         77.217 &          80.578 &        61.746 &        69.991 &         73.001 &        0.9246 &        0.9269 &         0.9181 &        0.3296 &        0.2527 &         0.2164 &               0.595  &               0.6742 &                0.7017 \\
  & & \textbf{MLP}      &0.1902 &0.0529&0.0708 &     8.787 &          8.53  &           9.135 &         68.991 &         78.262 &          80.487 &        60.204 &        69.732 &         71.352 &        0.9071 &        0.9129 &         0.9025 &        0.3355 &        0.2453 &         0.2163 &               0.5716 &               0.6676 &                0.6862 \\
\cellcolor{white}  &\cellcolor{white}   & \textbf{AdvDeb}    &     0.1151 &     0.1399 &  0.0879  &   30.046 &        34.488 &         34.968 &        36.11  &        38.694 &         43.041 &         6.064 &         \textbf{4.206} &          \textbf{8.073} &        0.7016 &        0.6668 &         0.6537 &        0.6427 &        0.6199 &         0.5812 &               0.0589 &               \textbf{0.0469} &                \textbf{0.0725} \\
 & & \textbf{lferm}     &     0.1448 &     \textbf{0.0194} &   \textbf{0.0386} &   31.459 &        28.632 &         24.965 &        31.764 &        47.464 &         57.47  &         \textbf{0.305} &        18.832 &         32.505 &        0.6857 &        0.7062 &         0.7314 &        0.6864 &        0.5632 &         0.4701 &               \textbf{0.0007} &               0.143  &                0.2613 \\

\multirow{-10}{*}{\cellcolor{white}Adult}   &\multirow{-10}{*}{\cellcolor{white}\textit{gender}}   & \textbf{FairC}    &\textbf{0.0528}&0.2451 &0.1274&    58.841 &         60.464 &          56.68  &         17.891 &         22.141 &          27.338 &        40.95  &        38.323 &         29.342 &        0.4135 &        0.3981 &         0.4219 &        0.8238 &        0.789  &         0.7415 &               0.4103 &               0.3909 &                0.3196 \\

  \midrule
   & & \textbf{LR}        &     0.1567 &     0.0695 &  0.0693   &    8.438 &        10.838 &         13.192 &        54.816 &        57.521 &         57.047 &        46.378 &        46.683 &         43.855 &        0.9239 &        0.9012 &         0.8736 &        0.464  &        0.4332 &         0.4303 &               0.4599 &               0.468  &                0.4433 \\
  \cellcolor{white}   &\cellcolor{white}   & \textbf{DT}      &\textbf{0.0438} &0.0492 &0.0272 &       6.334 &         14.796 &          17.298 &         31.092 &         49.802 &          54.639 &        24.758 &        35.006 &         37.341 &        0.9451 &        0.8723 &         0.8398 &        0.7224 &        0.5539 &         0.4936 &               0.2227 &               0.3184 &                0.3462 \\
    & & \textbf{SVM}      &     0.0534 &     0.0353 &    0.0227  &  11.937 &        16.377 &         17.379 &        31.305 &        33.869 &         35.385 &        \textbf{19.368} &        \textbf{17.492} &         \textbf{18.006} &        0.8871 &        0.8468 &         0.8295 &        0.6661 &        0.6616 &         0.6449 &               \textbf{0.221}  &               \textbf{0.1852} &                \textbf{0.1846}\\
  \cellcolor{white}   &\cellcolor{white}  & \textbf{LGBM}       &     0.1093 &     0.0470 &  0.0356   &    4.596 &         9.384 &         12.817 &        66.779 &        74.088 &         73.366 &        62.183 &        64.704 &         60.549 &        0.958  &        0.9185 &         0.8818 &        0.3744 &        0.2879 &         0.2804 &               0.5836 &               0.6306 &                0.6014 \\
    & & \textbf{XGB}       &     0.1056 &     0.0400 &   0.0304  &    1.803 &         3.152 &          6.523 &        81.289 &        88.9   &         84.48  &        79.486 &        85.748 &         77.957 &        0.9804 &        0.9711 &         0.9386 &        0.2183 &        0.1378 &         0.1599 &               0.7621 &               0.8333 &                0.7787 \\
 \cellcolor{white}   &\cellcolor{white}  & \textbf{RF}       &0.1058 &0.0703 &0.0461 &       3.616 &          6.067 &           8.473 &         71.126 &         81.269 &          81.498 &        67.51  &        75.202 &         73.025 &        0.9652 &        0.9452 &         0.9186 &        0.31   &        0.2151 &         0.201  &               0.6552 &               0.7301 &                0.7176 \\
  & & \textbf{MLP}       &0.0863&\textbf{0.0173} &0.0188&       0.402 &          0.935 &           1.934 &         92.918 &         96.854 &          96.337 &        92.516 &        95.919 &         94.403 &        0.9951 &        0.9917 &         0.9766 &        0.0889 &        0.0452 &         0.0435 &               0.9062 &               0.9465 &                0.9331 \\
  \cellcolor{white}   & \cellcolor{white} & \textbf{AdvDeb}    &     0.0957 &     0.0326 &  0.0282   &   16.369 &        20.067 &         23.123 &        44.722 &        52.645 &         57.043 &        28.353 &        32.578 &         33.92  &        0.8493 &        0.8119 &         0.7787 &        0.5803 &        0.4998 &         0.4536 &               0.269  &               0.3121 &                0.3251 \\
  &  & \textbf{lferm}     &     0.0639 &     0.0179 &    \textbf{0.0186}   &  8.943 &        13.316 &         16.561 &        47.036 &        54.87  &         55.83  &        38.093 &        41.554 &         39.269 &        0.9248 &        0.8809 &         0.8452 &        0.5618 &        0.4791 &         0.4584 &               0.363  &               0.4018 &                0.3868 \\
\multirow{-10}{*}{\cellcolor{white}AdultDeb}  &\multirow{-10}{*}{\cellcolor{white}\textit{gender}}  & \textbf{FairC}    &0.0575 &0.0529 &0.0315 &       1.326 &          2.723 &           4.359 &         80.127 &         85.728 &          88.23  &        78.801 &        83.005 &         83.871 &        0.9864 &        0.976  &         0.9556 &        0.1921 &        0.1533 &         0.1293 &               0.7943 &               0.8227 &                0.8263 \\
\midrule
    & & \textbf{LR}        &     0.6535 &     0.3294 &  0.3438  &     2.857 &         3.429 &          3.667 &        75.286 &        81.943 &         85.143 &        72.429 &        78.514 &         81.476 &        0.9688 &        0.9656 &         0.9568 &        0.2659 &        0.2011 &         0.1678 &               0.7029 &               0.7645 &                0.789  \\
 \cellcolor{white}     &\cellcolor{white}     & \textbf{DT}        &0.6190 &0.4039 &0.3827 &         7.2   &          6     &           6.32  &         65.211 &         75.239 &          79.254 &        58.011 &        69.239 &         72.934 &        0.9258 &        0.9376 &         0.9289 &        0.3648 &        0.2738 &         0.2321 &               0.561  &               0.6638 &                0.6968 \\
  &  & \textbf{SVM}    &     0.6395 &     0.3843 &  0.3390  &     6.25  &         5.917 &          5.63  &        73.239 &        80.789 &         84.493 &        66.989 &        74.872 &         78.863 &        0.938  &        0.94   &         0.9359 &        0.2868 &        0.2149 &         0.1776 &               0.6512 &               0.7251 &                0.7583 \\
   \cellcolor{white}     & \cellcolor{white}& \textbf{LGBM}    &     0.6363 &     0.2824 &  0.3525  &     5.652 &         5.913 &          5.696 &        74.571 &        80.143 &         83.55  &        68.919 &        74.23  &         77.854 &        0.9424 &        0.9407 &         0.9357 &        0.2875 &        0.2215 &         0.1854 &               0.6549 &               0.7192 &                0.7503 \\
 & & \textbf{XGB}    &     0.6568 &     0.2941 &   0.3656   &   5     &         5.455 &          5.045 &        73.38  &        80.141 &         83.613 &        68.38  &        74.686 &         78.568 &        0.9492 &        0.9467 &         0.9427 &        0.2938 &        0.2214 &         0.1851 &               0.6554 &               0.7253 &                0.7576 \\
 \cellcolor{white}     &\cellcolor{white}   & \textbf{RF}        &0.6363 &0.2824 &0.3525 &         8.261 &          7.478 &           7.652 &         65.417 &         73.972 &          77.918 &        57.156 &        66.494 &         70.266 &        0.9181 &        0.9245 &         0.9171 &        0.3813 &        0.2907 &         0.2477 &               0.5368 &               0.6338 &                0.6694 \\
  & & \textbf{MLP}    &0.6622 &0.3264 &0.3599 &         5.263 &          6.211 &           6.053 &         73.562 &         79.014 &          82.452 &        68.299 &        72.803 &         76.399 &        0.9509 &        0.9401 &         0.9337 &        0.2899 &        0.231  &         0.1953 &               0.661  &               0.7091 &                0.7384 \\
  
  \cellcolor{white}     &\cellcolor{white}   & \textbf{AdvDeb}   &     0.5501 &     0.1882 &  0.2732   &    7.576 &         7.03  &          7.273 &        69     &        77.571 &         80.643 &        61.424 &        70.541 &         73.37  &        0.9295 &        0.9295 &         0.9209 &        0.3396 &        0.2519 &         0.2164 &               0.5899 &               0.6776 &                0.7045 \\
 & &\textbf{lferm}   &     0.6125 &     0.2941 &   0.3278   &   3.913 &         6.174 &          6.696 &        64.412 &        71.588 &         75.103 &        60.499 &      65.414 &         68.407 &        0.9592 &        0.9408 &         0.929  &        0.3695 &        0.305  &         0.2686 &               0.5897 &               0.6358 &               0.6604 \\

 \multirow{-10}{*}{\cellcolor{white} Crime}  &\multirow{-10}{*}{\cellcolor{white}\textit{race}}  & \textbf{FairC}   &\textbf{0.2258} &\textbf{0.1373} &\textbf{0.0862} &        23.256 &         22.093 &          22.289 &         30.769 &         41.731 &          49.218 &         \textbf{7.513} &        \textbf{19.638} &         \textbf{26.929} &        0.7654 &        0.7742 &         0.7689 &        0.7248 &        0.6189 &         0.5455 &               \textbf{0.0406} &               \textbf{0.1553} &                \textbf{0.2234} \\

\midrule

 && \textbf{LR}      &     0.1187 &     0.1400 &  0.1657   &  31.364 &         35.455 &          39.081 & 27.500         & 27.500         & 27.250          &         3.864 &         7.955 &         11.831 &        0.6977 &        0.6582 &         0.6228 & 0.7302        & 0.7306        & 0.7267         &               0.0325 &               \textbf{0.0724} &                0.1039 \\
  \cellcolor{white}   &\cellcolor{white}   & \textbf{DT}       &\textbf{0.0271} &0.0500 & 0.0537 &     27.273 &         26.727 &          29.455 & 33.333         & 48.667         & 51.333          &         6.06  &        21.94  &         21.878 &        0.7486 &        0.7391 &         0.71   & 0.7246        & 0.5633        & 0.5217         &               0.024  &               0.1758 &                0.1883 \\
  && \textbf{SVM}    &     0.0449 &     0.0300 &  0.0892 &    22.5   &         31     &          32     & 30.000         & 70.000         & 70.000          &        7.5   &        39     &         38     &        0.8009 &        0.7185 &         0.6944 & 0.7722        & 0.4044        & 0.3569         &               0.0287 &               0.3141 &                0.3375 \\
 \cellcolor{white}   &\cellcolor{white}   & \textbf{LGBM}    &     0.1632 &     0.1900 &   0.11157   &      3.333 &          4.333 &           4.25  & 17.500         & 16.500         & 17.000          &        14.167 &        12.167 &         12.75  &        0.9706 &        0.9598 &         0.9528 & 0.8383        & 0.8379        & 0.8273         &               0.1323 &               0.1219 &                0.1255 \\
 && \textbf{XGB}   &     0.0518 &     0.0400 &   \textbf{0.0296} &    30     &         35.091 &          36.727 & 26.667         & 25.000         & 28.667          &         \textbf{3.333} &        10.091 &          \textbf{8.06}  &        0.6891 &        0.6563 &         0.6364 & 0.7945        & 0.7686        & 0.7271         &               0.1054 &               0.1123 &                \textbf{0.0907} \\
\cellcolor{white}   &\cellcolor{white}   & \textbf{RF}       &0.0626 &0.0800 &0.0878 &     32     &         33.2   &          35.046 & 40.000         & 46.000         & 47.303          &         8     &        12.8   &         12.257 &        0.6957 &        0.6751 &         0.6537 & 0.6779        & 0.5816        & 0.5530         &               \textbf{0.0178} &               0.0935 &                0.1007 \\
 & & \textbf{MLP}   &0.0449 &0.0300 &0.0892 &     23     &         29.4   &          30.097 & $0^*$            & $0^*$            & $0^*$             &        23     &        29.4   &         30.097 &        0.7726 &        0.7184 &         0.7039 & $0^*$           & $0^*$           & $0^*$            &               0.7726 &               0.7184 &                0.7039 \\
  \cellcolor{white}   &\cellcolor{white}  & \textbf{AdvDeb}   &     0.1809 &     0.2200 &  0.1267     &       5     &          7.667 &           6.583 & 15.714         & 14.857         & 17.286          &        10.714 &         \textbf{7.19}  &         10.703 &        0.9573 &        0.9315 &         0.9295 & 0.8299        & 0.8458        & 0.8243         &               0.1274 &               0.0857 &                0.1052 \\
  & & \textbf{lferm}   &     0.0355 &     \textbf{0.0200} &   0.0746   &     15     &         19     &          24     & $0^*$            & $0^*$            & $0^*$             &        15     &        19     &         24     &        0.8408 &        0.8156 &         0.7676 & $0^*$           & $0^*$           & $0^*$            &               0.8408 &               0.8156 &                0.7676 \\
\multirow{-10}{*}{\cellcolor{white}German} &\multirow{-10}{*}{\cellcolor{white}\textit{gender}} & \textbf{FairC}    &0.0449 &0.0500 &0.0728 &     24     &         27     &          29.019 & 0.000          & 4.000          & 5.000           &        24     &        23     &         24.019 &        0.7679 &        0.734  &         0.7124 & 1.0000        & 0.9705        & 0.9505         &               0.2321 &               0.2365 &                0.2381 \\

\bottomrule
\hline
\end{tabular}
\end{adjustbox}
\end{table*}

%% file: tables/CFlipsKDtreeNeg.tex
\begin{table*}[ht]
    \centering
    \scriptsize
    \caption{(KDtree) \text{CFlip} and \text{nDCCF} results at different $|k|$ number of Counterfactuals for each negatively predicted Test set sample, ($0^*$ there are no negative predicted \textit{unprivileged} samples which result in no CF samples for the unprivileged group). We mark the best-performing model for each fairness metric in bold font.}
    \label{tab:totKDtreeCFLIPNDCCF}
    \begin{adjustbox}{width=\textwidth, center}
\setlength{\tabcolsep}{2.5pt}
\renewcommand{\arraystretch}{1}
\rowcolors{12}{gray!15}{white}
\begin{tabular}{lllrrr|rrr|rrr|rrr|rrr|rrr}
\hline
\toprule
&& & \multicolumn{9}{c}{$\text{CFlips}$@$|k|$ (\%)} & \multicolumn{9}{c}{$\text{nDCCF}@|k|$}
\\ \cmidrule(lr){4-12} \cmidrule(lr){13-21}
& &&  \multicolumn{3}{c}{Privileged} & \multicolumn{3}{c}{Unprivileged} & \multicolumn{3}{c}{$\Delta\text{CFlips}\downarrow
$}&\multicolumn{3}{c}{Privileged} & \multicolumn{3}{c}{Unprivileged} & \multicolumn{3}{c}{$\Delta\text{nDCCF}\downarrow
$} \\ \cmidrule(lr){4-6} \cmidrule(lr){7-9} \cmidrule(lr){10-12} \cmidrule(lr){13-15} \cmidrule(lr){16-18} \cmidrule(lr){19-21}  
 Dataset   & $s$&model   &   @10 &   @50 &   \multicolumn{1}{c}{@100} &   @10 &   @50 &   \multicolumn{1}{c}{@100} &   @10 &   @50 &   \multicolumn{1}{c}{@100} &   @10 & @50 & \multicolumn{1}{c}{@100} & @10 &   @50 &   \multicolumn{1}{c}{@100} & @10 & @50 &   @100 \\
\midrule
 & &\textbf{LR}      &          7.754 &          8.584 &           9.11  &         81.273 &         84.21  &          85.137 &        73.519 &        75.626 &         76.027 &        0.9229 &        0.9162 &         0.9049 &        0.2053 &        0.1701 &         0.1573 &               0.7176 &               0.7461 &                0.7476 \\
\cellcolor{white}   &\cellcolor{white}   & \textbf{DT}      &          6.971 &          6.927 &           7.371 &         76.419 &         82.602 &          84.523 &        69.448 &        75.675 &         77.152 &        0.9296 &        0.9305 &         0.9207 &        0.2567 &        0.1941 &         0.1707 &               0.6729 &               0.7364 &                0.75   \\
&  & \textbf{SVM}     &          6.306 &          6.427 &           7.158 &         80.971 &         85.209 &          86.517 &        74.665 &        78.782 &         79.359 &        0.9343 &        0.9351 &         0.9232 &        0.2088 &        0.1632 &         0.1465 &               0.7255 &               0.7719 &                0.7767 \\
\cellcolor{white}   &\cellcolor{white}   & \textbf{LGBM}     &          6.55  &          7.077 &           7.663 &         82.009 &         84.902 &          86.064 &        75.459 &        77.825 &         78.401 &        0.9352 &        0.9307 &         0.9192 &        0.1889 &        0.1602 &         0.1469 &               0.7463 &               0.7705 &                0.7723 \\
  && \textbf{XGB}     &          6.761 &          7.14  &           7.666 &         81.677 &         84.796 &          86.076 &        74.916 &        77.656 &         78.41  &        0.9334 &        0.9299 &         0.9189 &        0.192  &        0.1616 &         0.1472 &               0.7414 &               0.7683 &                0.7717 \\
 \cellcolor{white}   &\cellcolor{white}  & \textbf{RF}      &          6.44  &          7.578 &           8.076 &         80.218 &         83.325 &          84.767 &        73.778 &        75.747 &         76.691 &        0.9362 &        0.9268 &         0.9155 &        0.2161 &        0.1795 &         0.1627 &               0.7201 &               0.7473 &                0.7528 \\
&  & \textbf{MLP}     &          7.153 &          7.464 &           7.94  &         81.455 &         84.484 &          85.888 &        74.302 &        77.02  &         77.948 &        0.93   &        0.9266 &         0.9161 &        0.1965 &        0.1654 &         0.1498 &               0.7335 &               0.7612 &                0.7663 \\
 \cellcolor{white}   &\cellcolor{white}  & \textbf{AdvDeb}  &         25.411 &         18.854 &          17.847 &         52.22  &         66.941 &          71.783 &        26.809 &        48.087 &         53.936 &        0.7368 &        0.7946 &         0.8029 &        0.5015 &        0.37   &         0.3159 &               0.2353 &               0.4246 &                0.487  \\
  && \textbf{LFERM}   &         15.141 &         11.82  &          11.882 &         66.288 &         77.278 &          80.16  &        51.147 &        65.458 &         68.278 &        0.8385 &        0.8715 &         0.8689 &        0.3719 &        0.262  &         0.2255 &               0.4666 &               0.6095 &                0.6434 \\
\multirow{-10}{*}{\cellcolor{white}Adult}    &\multirow{-10}{*}{\cellcolor{white}\textit{gender}}    &  \textbf{FairC}   &         37.337 &         27.402 &          25.323 &         37.832 &         55.442 &          61.712 &         \textbf{0.495} &        \textbf{28.04}  &         \textbf{36.389} &        0.6137 &        0.7005 &         0.7208 &        0.6442 &        0.4916 &         0.4238 &               \textbf{0.0305} &               \textbf{0.2089} &                \textbf{0.297}  \\
\midrule
 && \textbf{LR}      &         17.578 &         19.789 &          20.026 &         42.592 &         44.051 &          45.165 &        25.014 &        24.262 &         25.139 &        0.8422 &        0.8066 &         0.8018 &        0.5742 &        0.5612 &         0.5473 &               0.268  &               0.2454 &                0.2545 \\
 \cellcolor{white}&\cellcolor{white}& \textbf{DT}      &         25.93  &         42.627 &          41.962 &         31.724 &         44.805 &          55.245 &         \textbf{5.794} &         \textbf{2.178} &         \textbf{13.283} &        0.779  &        0.6248 &         0.6037 &        0.6637 &        0.5775 &         0.4877 &               \textbf{0.1153} &               \textbf{0.0473} &                \textbf{0.116}  \\
 & & \textbf{SVM}     &         15.283 &         16.551 &          16.574 &         42.549 &         43.343 &          44.088 &        27.266 &        26.792 &         27.514 &        0.8427 &        0.8211 &         0.8184 &        0.5705 &        0.5621 &         0.5524 &               0.2722 &               0.259  &                0.266  \\
\cellcolor{white}&\cellcolor{white} & \textbf{LGBM}     &          9.811 &         14.834 &          15.608 &         45.253 &         69.258 &          75.566 &        35.442 &        54.424 &         59.958 &        0.907  &        0.8626 &         0.8457 &        0.568  &        0.3686 &         0.2978 &               0.339  &               0.494  &                0.5479 \\
 && \textbf{XGB}     &          8.955 &         12.959 &          14.183 &         49.229 &         70.872 &          76.837 &        40.274 &        57.913 &         62.654 &        0.9149 &        0.88   &         0.8601 &        0.5303 &        0.3479 &         0.2805 &               0.3846 &               0.5321 &                0.5796 \\
\cellcolor{white}&\cellcolor{white} & \textbf{RF}      &         10.354 &         14.39  &          15.675 &         45.517 &         65.841 &          73.207 &        35.163 &        51.451 &         57.532 &        0.9024 &        0.8645 &         0.8469 &        0.5564 &        0.3953 &         0.3239 &               0.346  &               0.4692 &                0.523  \\
 & &\textbf{MLP}     &          5.241 &          8.797 &           7.754 &         71.167 &         85.204 &          89.1   &        65.926 &        76.407 &         81.346 &        0.9414 &        0.917  &         0.9141 &        0.3445 &        0.194  &         0.1534 &               0.5969 &               0.723  &                0.7607 \\
\cellcolor{white}&\cellcolor{white} & \textbf{AdvDeb}  &          8.608 &         14.501 &          13.277 &         51.118 &         72.431 &          78.922 &        42.51  &        57.93  &         65.645 &        0.917  &        0.8695 &         0.8663 &        0.4993 &        0.3248 &         0.2613 &               0.4177 &               0.5447 &                0.605  \\
  \cellcolor{white}&\cellcolor{white}& \textbf{LFERM}   &          7.028 &         10.736 &          13.216 &         65.974 &         73.594 &          73.763 &        58.946 &        62.858 &         60.547 &        0.933  &        0.9022 &         0.8737 &        0.3571 &        0.2864 &         0.2755 &               0.5759 &               0.6158 &                0.5982 \\
  
\multirow{-10}{*}{\cellcolor{white}AdultDeb} &\multirow{-10}{*}{\cellcolor{white}\textit{gender}} & \textbf{FairC}   &          7.191 &         12.942 &          11.435 &         64.482 &         78.892 &          85.475 &        57.291 &        65.95  &         74.04  &        0.9266 &        0.8833 &         0.8827 &        0.4124 &        0.2565 &         0.1892 &               0.5142 &               0.6268 &                0.6935 \\

\midrule
 & &\textbf{LR}      &          5.714 &          8.762 &           9.143 &         85.139 &         87.722 &          88     &        79.425 &        78.96  &         78.857 &        0.9489 &        0.9207 &         0.9077 &        0.1603 &        0.1319 &         0.1257 &               0.7886 &               0.7888 &                0.782  \\
 \cellcolor{white} &\cellcolor{white} & \textbf{\textbf{DT}}      &         10.4   &         12.96  &          14.6   &         75.07  &         80.113 &          80.789 &        64.67  &        67.153 &         66.189 &        0.8765 &        0.8695 &         0.8512 &        0.258  &        0.2126 &         0.2007 &               0.6185 &               0.6569 &                0.6505 \\
  && \textbf{SVM}     &          9.167 &         10.667 &          11.75  &         82.877 &         86.466 &          86.959 &        73.71  &        75.799 &         75.209 &        0.9011 &        0.8937 &         0.8788 &        0.1716 &        0.1424 &         0.1351 &               0.7295 &               0.7513 &                0.7437 \\
 \cellcolor{white} &\cellcolor{white} & \textbf{LGBM}     &          4.348 &          8.522 &           9.783 &         81.806 &         85.444 &          86.236 &        77.458 &        76.922 &         76.453 &        0.9528 &        0.9221 &         0.9033 &        0.1892 &        0.1551 &         0.1443 &               0.7636 &               0.767  &                0.759  \\
  && \textbf{XGB}     &          5.455 &          8.727 &           9.727 &         83.288 &         86.329 &          86.63  &        77.833 &        77.602 &         76.903 &        0.9308 &        0.9149 &         0.8992 &        0.1767 &        0.1463 &         0.1396 &               0.7541 &               0.7686 &                0.7596 \\
 \cellcolor{white} &\cellcolor{white} & \textbf{RF}      &         10.87  &         13.043 &          14.261 &         78.75  &         81.694 &          82.486 &        67.88  &        68.651 &         68.225 &        0.8861 &        0.8736 &         0.857  &        0.2214 &        0.1926 &         0.1822 &               0.6647 &               0.681  &                0.6748 \\
  & &\textbf{MLP}     &          4.5   &          8.4   &          10     &         80.972 &         84.389 &          85.347 &        76.472 &        75.989 &         75.347 &        0.9611 &        0.926  &         0.9028 &        0.194  &        0.164  &         0.1526 &               0.7671 &               0.762  &                0.7502 \\
 \cellcolor{white} &\cellcolor{white} & \textbf{AdvDeb}  &          8.485 &          9.818 &          10.667 &         81.389 &         83.889 &          84.861 &        72.904 &        74.071 &         74.194 &        0.9134 &        0.9034 &         0.8895 &        0.1912 &        0.1685 &         0.1577 &               0.7222 &               0.7349 &                0.7318 \\
  && \textbf{LFERM}   &          9.13  &         14     &          15.174 &         71.857 &         77.971 &          78.886 &        62.727 &        63.971 &         63.712 &        0.9119 &        0.8709 &         0.8516 &        0.291  &        0.2371 &         0.2224 &               0.6209 &               0.6338 &                0.6292 \\
  
 \multirow{-10}{*}{\cellcolor{white} Crime}   &\multirow{-10}{*}{\cellcolor{white}\textit{race}}   & \textbf{FairC}   &         33.023 &         32.512 &          34.186 &         39.808 &         48.885 &          52.577 &         \textbf{6.785} &        \textbf{16.373} &         \textbf{18.391} &        0.6808 &        0.6787 &         0.6607 &        0.6207 &        0.5395 &         0.4981 &               \textbf{0.0601} &               \textbf{0.1392} &                \textbf{0.1626} \\
\midrule
& & \textbf{LR}      &         23.182 &         26.545 &          26.818 & 45.000         & 56.000         & 62.000          &        21.818 &        29.455 &         35.182 &        0.761  &        0.7407 &         0.732  & 0.5683        & 0.4618        & 0.4006         &               0.1927 &               0.2789 &                0.3314 \\
\cellcolor{white} &\cellcolor{white} & \textbf{DT}      &         28.182 &         27.818 &          28.273 & 60.000         & 57.333         & 61.333          &        31.818 &        29.515 &         33.06  &        0.7362 &        0.7272 &         0.7174 & 0.4351        & 0.4332        & 0.3958         &               0.3011 &               0.294  &                0.3216 \\
 & &\textbf{SVM}     &         23.333 &         26.333 &          27.417 & 40.000         & 56.000         & 65.000          &        16.667 &        29.667 &         37.583 &        0.7731 &        0.7459 &         0.7295 & 0.6805        & 0.5024        & 0.4122         &               0.0926 &               0.2435 &                0.3173 \\
\cellcolor{white} &\cellcolor{white} & \textbf{LGBM}     &         26.667 &         29.167 &          29.417 & 37.500         & 50.000         & 58.250          &        10.833 &        20.833 &         28.833 &        0.7616 &        0.7249 &         0.7124 & 0.6265        & 0.5235        & 0.4519         &               0.1351 &               0.2014 &                0.2605 \\
 && \textbf{XGB}     &         22.727 &         27.818 &          27.909 & 46.667         & 53.333         & 60.167          &        23.94  &        25.515 &         32.258 &        0.7976 &        0.7424 &         0.7301 & 0.5489        & 0.4807        & 0.4211         &               0.2487 &               0.2617 &                0.309  \\
\cellcolor{white} &\cellcolor{white} & \textbf{RF}      &         23     &         27.6   &          26.6   & 55.000         & 58.000         & 64.000          &        32     &        30.4   &         37.4   &        0.7951 &        0.7414 &         0.7379 & 0.5127        & 0.4424        & 0.3874         &               0.2824 &               0.299  &                0.3505 \\
 & & \textbf{MLP}     &         30     &         29.6   &          30     & $0^*$            & $0^*$            & $0^*$             &        30     &        29.6   &         30     &        0.726  &        0.7145 &         0.7024 & $0^*$           & $0^*$           & $0^*$            &               0.726  &               0.7145 &                0.7024 \\
 \cellcolor{white} &\cellcolor{white}& \textbf{AdvDeb}  &         28.333 &         30     &          29.833 & 40.000         & 56.286         & 61.429          &        11.667 &        26.286 &         31.596 &        0.74   &        0.7128 &         0.7027 & 0.6109        & 0.4703        & 0.4155         &               0.1291 &               0.2425 &                0.2872 \\
 && \textbf{LFERM}   &         20     &         23     &          23.5   & $0^*$            & $0^*$            & $0^*$             &        20     &        23     &         23.5   &        0.7702 &        0.7609 &         0.7567 & $0^*$           & $0^*$           & $0^*$            &               0.7702 &               0.7609 &                0.7567 \\
  \multirow{-10}{*}{\cellcolor{white} German} &\multirow{-10}{*}{\cellcolor{white} \textit{gender}} & \textbf{FairC}   &         27     &         29.4   &          28.7   & 20.000         & 50.000         & 60.000          &         \textbf{7}     &        \textbf{20.6}   &         \textbf{31.3}   &        0.745  &        0.7174 &         0.7127 & 0.7760        & 0.5517        & 0.4571         &               \textbf{0.031}  &               \textbf{0.1657} &                \textbf{0.2556} \\

\bottomrule
\hline
\end{tabular}
\end{adjustbox}
\end{table*}

%% file: groupplot/proxyfeature.tex
\begin{figure}[!t]
\tiny
\centering
\includegraphics[width=\textwidth]{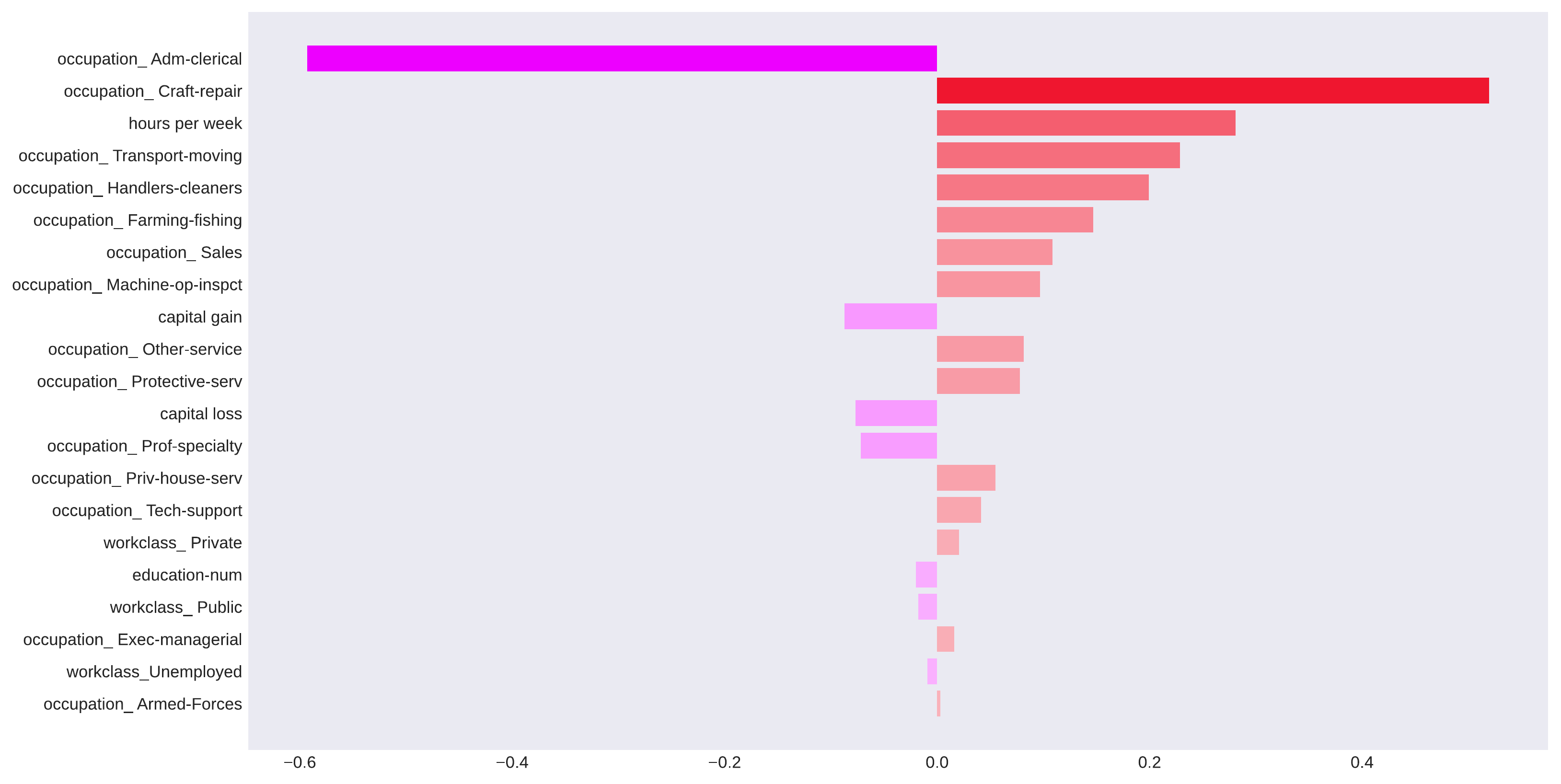}
\caption{Features correlation rank with a \textit{gender} Flip (i.e., $\rho(\mathcal{E},\Delta)$ on the x-axis) on the Adult-debiased dataset with Genetic strategy as $g(\cdot)$, MLP as $f(\cdot)$ and XGB $f_s(\cdot)$ for only sample, and, thus, counterfactuals, belonging to $\mathcal{X}^-$.
}
    \label{fig:explainCorr}
\end{figure}

%% file: groupplot/ablationPlots.tex
\begin{figure}[!t]
\pgfplotsset{tiny,width=5.3cm,compat=1.17}
\centering
\begin{subfigure}{1\textwidth}
\centering
\begin{tikzpicture}
	\begin{axis}[
	legend columns = -1,
	legend entries = {LR, DT, SVM, LGBM, XGB, RF, MLP, LFERM, AdvDeb, FairC},
	legend to name = name,
    title={RF},
    yticklabel={\pgfmathprintnumber\tick\%},
    ylabel={$\Delta\mathrm{CFlips}(\mathcal{X}^{-})$},
    xmin=-1, xmax=100, height=4cm, 
    ymin=0, ymax=100,
]
\addplot[color=clrLR,
    mark=none,] table [x=i, y=j, col sep=comma ] {csv/CFlips/RF/LR.csv};
\addplot[color=clrDT,
    mark=none,] table [x=i, y=j, col sep=comma ] {csv/CFlips/RF/DT.csv};
\addplot[color=clrSVM,
    mark=none,] table [x=i, y=j, col sep=comma ] {csv/CFlips/RF/SVM.csv};
\addplot[color=clrLGB,
    mark=none,] table [x=i, y=j, col sep=comma ] {csv/CFlips/RF/LGB.csv};
\addplot[color=clrXGB,
    mark=none,] table [x=i, y=j, col sep=comma ] {csv/CFlips/RF/XGB.csv};
\addplot[color=clRF,
    mark=none,] table [x=i, y=j, col sep=comma ] {csv/CFlips/RF/RF.csv};
\addplot[color=clMLP,
    mark=none,] table [x=i, y=j, col sep=comma ] {csv/CFlips/RF/MLP.csv};
\addplot[color=clLFERM,
    mark=none,] table [x=i, y=j, col sep=comma ] {csv/CFlips/RF/lferm.csv};
\addplot[color=clADV,
    mark=none,] table [x=i, y=j, col sep=comma ] {csv/CFlips/RF/Adv.csv};
\addplot[color=clZafar,
    mark=none,] table [x=i, y=j, col sep=comma ] {csv/CFlips/RF/zafar.csv};
    \end{axis}
\end{tikzpicture}
\begin{tikzpicture}
	\begin{axis}[
    title={MLP},
    yticklabel={\empty},
    xmin=-1, xmax=100, height=4cm, 
    ymin=0, ymax=100,
]
\addplot[color=clrLR,
    mark=none,] table [x=i, y=j, col sep=comma ] {csv/CFlips/MLP/LR.csv};
\addplot[color=clrDT,
    mark=none,] table [x=i, y=j, col sep=comma ] {csv/CFlips/MLP/DT.csv};
\addplot[color=clrSVM,
    mark=none,] table [x=i, y=j, col sep=comma ] {csv/CFlips/MLP/SVM.csv};
\addplot[color=clrLGB,
    mark=none,] table [x=i, y=j, col sep=comma ] {csv/CFlips/MLP/LGB.csv};
\addplot[color=clrXGB,
    mark=none,] table [x=i, y=j, col sep=comma ] {csv/CFlips/MLP/XGB.csv};
\addplot[color=clRF,
    mark=none,] table [x=i, y=j, col sep=comma ] {csv/CFlips/MLP/RF.csv};
\addplot[color=clMLP,
    mark=none,] table [x=i, y=j, col sep=comma ] {csv/CFlips/MLP/MLP.csv};
\addplot[color=clLFERM,
    mark=none,] table [x=i, y=j, col sep=comma ] {csv/CFlips/MLP/lferm.csv};
\addplot[color=clADV,
    mark=none,] table [x=i, y=j, col sep=comma ] {csv/CFlips/MLP/Adv.csv};
\addplot[color=clZafar,
    mark=none,] table [x=i, y=j, col sep=comma ] {csv/CFlips/MLP/zafar.csv};
    \end{axis}
\end{tikzpicture}
\begin{tikzpicture}
	\begin{axis}[
    title={XGB},
    yticklabel={\empty},
    xmin=-1, xmax=100, height=4cm, 
    ymin=0, ymax=100,
]
\addplot[color=clrLR,
    mark=none,] table [x=i, y=j, col sep=comma ] {csv/CFlips/XGB/LR.csv};
\addplot[color=clrDT,
    mark=none,] table [x=i, y=j, col sep=comma ] {csv/CFlips/XGB/DT.csv};
\addplot[color=clrSVM,
    mark=none,] table [x=i, y=j, col sep=comma ] {csv/CFlips/XGB/SVM.csv};
\addplot[color=clrLGB,
    mark=none,] table [x=i, y=j, col sep=comma ] {csv/CFlips/XGB/LGB.csv};
\addplot[color=clrXGB,
    mark=none,] table [x=i, y=j, col sep=comma ] {csv/CFlips/XGB/XGB.csv};
\addplot[color=clRF,
    mark=none,] table [x=i, y=j, col sep=comma ] {csv/CFlips/XGB/RF.csv};
\addplot[color=clMLP,
    mark=none,] table [x=i, y=j, col sep=comma ] {csv/CFlips/XGB/MLP.csv};
\addplot[color=clLFERM,
    mark=none,] table [x=i, y=j, col sep=comma ] {csv/CFlips/XGB/lferm.csv};
\addplot[color=clADV,
    mark=none,] table [x=i, y=j, col sep=comma ] {csv/CFlips/XGB/Adv.csv};
\addplot[color=clZafar,
    mark=none,] table [x=i, y=j, col sep=comma ] {csv/CFlips/XGB/zafar.csv};
    \end{axis}
\end{tikzpicture}
\end{subfigure}
\begin{subfigure}{1\textwidth}
\centering
\begin{tikzpicture}
	\begin{axis}[
	legend columns = -1,
	legend entries = {LR, DT, SVM, LGBM, XGB, RF, MLP, LFERM, AdvDeb, FairC},
	legend to name = name,
    title={RF},
    yticklabel={\pgfmathprintnumber\tick},
    ylabel={$\Delta\mathrm{nDCCF}(\mathcal{X}^{-})$},
    xmin=-1, xmax=100, height=4cm, 
    ymin=0, ymax=1,
]
\addplot[color=clrLR,
    mark=none,] table [x=i, y=j, col sep=comma ] {csv/nDCCF/RF/LR.csv};
\addplot[color=clrDT,
    mark=none,] table [x=i, y=j, col sep=comma ] {csv/nDCCF/RF/DT.csv};
\addplot[color=clrSVM,
    mark=none,] table [x=i, y=j, col sep=comma ] {csv/nDCCF/RF/SVM.csv};
\addplot[color=clrLGB,
    mark=none,] table [x=i, y=j, col sep=comma ] {csv/nDCCF/RF/LGB.csv};
\addplot[color=clrXGB,
    mark=none,] table [x=i, y=j, col sep=comma ] {csv/nDCCF/RF/XGB.csv};
\addplot[color=clRF,
    mark=none,] table [x=i, y=j, col sep=comma ] {csv/nDCCF/RF/RF.csv};
\addplot[color=clMLP,
    mark=none,] table [x=i, y=j, col sep=comma ] {csv/nDCCF/RF/MLP.csv};
\addplot[color=clLFERM,
    mark=none,] table [x=i, y=j, col sep=comma ] {csv/nDCCF/RF/lferm.csv};
\addplot[color=clADV,
    mark=none,] table [x=i, y=j, col sep=comma ] {csv/nDCCF/RF/Adv.csv};
\addplot[color=clZafar,
    mark=none,] table [x=i, y=j, col sep=comma ] {csv/nDCCF/RF/zafar.csv};
    \end{axis}
\end{tikzpicture}
\begin{tikzpicture}
	\begin{axis}[
    title={MLP},
    yticklabel={\empty},
    xmin=-1, xmax=100, height=4cm, 
    ymin=0, ymax=1,
]
\addplot[color=clrLR,
    mark=none,] table [x=i, y=j, col sep=comma ] {csv/nDCCF/MLP/LR.csv};
\addplot[color=clrDT,
    mark=none,] table [x=i, y=j, col sep=comma ] {csv/nDCCF/MLP/DT.csv};
\addplot[color=clrSVM,
    mark=none,] table [x=i, y=j, col sep=comma ] {csv/nDCCF/MLP/SVM.csv};
\addplot[color=clrLGB,
    mark=none,] table [x=i, y=j, col sep=comma ] {csv/nDCCF/MLP/LGB.csv};
\addplot[color=clrXGB,
    mark=none,] table [x=i, y=j, col sep=comma ] {csv/nDCCF/MLP/XGB.csv};
\addplot[color=clRF,
    mark=none,] table [x=i, y=j, col sep=comma ] {csv/nDCCF/MLP/RF.csv};
\addplot[color=clMLP,
    mark=none,] table [x=i, y=j, col sep=comma ] {csv/nDCCF/MLP/MLP.csv};
\addplot[color=clLFERM,
    mark=none,] table [x=i, y=j, col sep=comma ] {csv/nDCCF/MLP/lferm.csv};
\addplot[color=clADV,
    mark=none,] table [x=i, y=j, col sep=comma ] {csv/nDCCF/MLP/Adv.csv};
\addplot[color=clZafar,
    mark=none,] table [x=i, y=j, col sep=comma ] {csv/nDCCF/MLP/zafar.csv};
    \end{axis}
\end{tikzpicture}
\begin{tikzpicture}
	\begin{axis}[
    title={XGB},
    yticklabel={\empty},
    xmin=-1, xmax=100, height=4cm, 
    ymin=0, ymax=1,
]
\addplot[color=clrLR,
    mark=none,] table [x=i, y=j, col sep=comma ] {csv/nDCCF/XGB/LR.csv};
\addplot[color=clrDT,
    mark=none,] table [x=i, y=j, col sep=comma ] {csv/nDCCF/XGB/DT.csv};
\addplot[color=clrSVM,
    mark=none,] table [x=i, y=j, col sep=comma ] {csv/nDCCF/XGB/SVM.csv};
\addplot[color=clrLGB,
    mark=none,] table [x=i, y=j, col sep=comma ] {csv/nDCCF/XGB/LGB.csv};
\addplot[color=clrXGB,
    mark=none,] table [x=i, y=j, col sep=comma ] {csv/nDCCF/XGB/XGB.csv};
\addplot[color=clRF,
    mark=none,] table [x=i, y=j, col sep=comma ] {csv/nDCCF/XGB/RF.csv};
\addplot[color=clMLP,
    mark=none,] table [x=i, y=j, col sep=comma ] {csv/nDCCF/XGB/MLP.csv};
\addplot[color=clLFERM,
    mark=none,] table [x=i, y=j, col sep=comma ] {csv/nDCCF/XGB/lferm.csv};
\addplot[color=clADV,
    mark=none,] table [x=i, y=j, col sep=comma ] {csv/nDCCF/XGB/Adv.csv};
\addplot[color=clZafar,
    mark=none,] table [x=i, y=j, col sep=comma ] {csv/nDCCF/XGB/zafar.csv};
    \end{axis}
\end{tikzpicture}
\end{subfigure}
\begin{subfigure}{1\textwidth}
\centering
\begin{tikzpicture}
	\begin{axis}[
	legend columns = -1,
	legend entries = {LR, DT, SVM, LGBM, XGB, RF, MLP, LFERM, AdvDeb, FairC},
	legend to name = name,
    title={RF},
    xlabel={$\mid \mathcal{C}_\mathbf{x} \mid$},
    yticklabel={\pgfmathprintnumber\tick},
    ylabel={$\Delta\mathrm{nDCCF}_{sorted}(\mathcal{X}^{-})$},
    xmin=-1, xmax=100, height=4cm, 
    ymin=0, ymax=1,
]
\addplot[color=clrLR,
    mark=none,] table [x=i, y=j, col sep=comma ] {csv/reRanknDCCF/RF/LR.csv};
\addplot[color=clrDT,
    mark=none,] table [x=i, y=j, col sep=comma ] {csv/reRanknDCCF/RF/DT.csv};
\addplot[color=clrSVM,
    mark=none,] table [x=i, y=j, col sep=comma ] {csv/reRanknDCCF/RF/SVM.csv};
\addplot[color=clrLGB,
    mark=none,] table [x=i, y=j, col sep=comma ] {csv/reRanknDCCF/RF/LGB.csv};
\addplot[color=clrXGB,
    mark=none,] table [x=i, y=j, col sep=comma ] {csv/reRanknDCCF/RF/XGB.csv};
\addplot[color=clRF,
    mark=none,] table [x=i, y=j, col sep=comma ] {csv/reRanknDCCF/RF/RF.csv};
\addplot[color=clMLP,
    mark=none,] table [x=i, y=j, col sep=comma ] {csv/reRanknDCCF/RF/MLP.csv};
\addplot[color=clLFERM,
    mark=none,] table [x=i, y=j, col sep=comma ] {csv/reRanknDCCF/RF/lferm.csv};
\addplot[color=clADV,
    mark=none,] table [x=i, y=j, col sep=comma ] {csv/reRanknDCCF/RF/Adv.csv};
\addplot[color=clZafar,
    mark=none,] table [x=i, y=j, col sep=comma ] {csv/reRanknDCCF/RF/zafar.csv};
    \end{axis}
\end{tikzpicture}
\begin{tikzpicture}
	\begin{axis}[
    title={MLP},
    yticklabel={\empty},
    xlabel={$\mid \mathcal{C}_\mathbf{x} \mid$},
    xmin=-1, xmax=100, height=4cm, 
    ymin=0, ymax=1,
]
\addplot[color=clrLR,
    mark=none,] table [x=i, y=j, col sep=comma ] {csv/reRanknDCCF/MLP/LR.csv};
\addplot[color=clrDT,
    mark=none,] table [x=i, y=j, col sep=comma ] {csv/reRanknDCCF/MLP/DT.csv};
\addplot[color=clrSVM,
    mark=none,] table [x=i, y=j, col sep=comma ] {csv/reRanknDCCF/MLP/SVM.csv};
\addplot[color=clrLGB,
    mark=none,] table [x=i, y=j, col sep=comma ] {csv/reRanknDCCF/MLP/LGB.csv};
\addplot[color=clrXGB,
    mark=none,] table [x=i, y=j, col sep=comma ] {csv/reRanknDCCF/MLP/XGB.csv};
\addplot[color=clRF,
    mark=none,] table [x=i, y=j, col sep=comma ] {csv/reRanknDCCF/MLP/RF.csv};
\addplot[color=clMLP,
    mark=none,] table [x=i, y=j, col sep=comma ] {csv/reRanknDCCF/MLP/MLP.csv};
\addplot[color=clLFERM,
    mark=none,] table [x=i, y=j, col sep=comma ] {csv/reRanknDCCF/MLP/lferm.csv};
\addplot[color=clADV,
    mark=none,] table [x=i, y=j, col sep=comma ] {csv/reRanknDCCF/MLP/Adv.csv};
\addplot[color=clZafar,
    mark=none,] table [x=i, y=j, col sep=comma ] {csv/reRanknDCCF/MLP/zafar.csv};
    \end{axis}
\end{tikzpicture}
\begin{tikzpicture}
	\begin{axis}[
    title={XGB},
    xlabel={$\mid \mathcal{C}_\mathbf{x} \mid$},
    yticklabel={\empty},
    xmin=-1, xmax=100, height=4cm, 
    ymin=0, ymax=1,
]
\addplot[color=clrLR,
    mark=none,] table [x=i, y=j, col sep=comma ] {csv/reRanknDCCF/XGB/LR.csv};
\addplot[color=clrDT,
    mark=none,] table [x=i, y=j, col sep=comma ] {csv/reRanknDCCF/XGB/DT.csv};
\addplot[color=clrSVM,
    mark=none,] table [x=i, y=j, col sep=comma ] {csv/reRanknDCCF/XGB/SVM.csv};
\addplot[color=clrLGB,
    mark=none,] table [x=i, y=j, col sep=comma ] {csv/reRanknDCCF/XGB/LGB.csv};
\addplot[color=clrXGB,
    mark=none,] table [x=i, y=j, col sep=comma ] {csv/reRanknDCCF/XGB/XGB.csv};
\addplot[color=clRF,
    mark=none,] table [x=i, y=j, col sep=comma ] {csv/reRanknDCCF/XGB/RF.csv};
\addplot[color=clMLP,
    mark=none,] table [x=i, y=j, col sep=comma ] {csv/reRanknDCCF/XGB/MLP.csv};
\addplot[color=clLFERM,
    mark=none,] table [x=i, y=j, col sep=comma ] {csv/reRanknDCCF/XGB/lferm.csv};
\addplot[color=clADV,
    mark=none,] table [x=i, y=j, col sep=comma ] {csv/reRanknDCCF/XGB/Adv.csv};
\addplot[color=clZafar,
    mark=none,] table [x=i, y=j, col sep=comma ] {csv/reRanknDCCF/XGB/zafar.csv};
    \end{axis}
\end{tikzpicture}
\end{subfigure}
\pgfplotsset{tiny,width=2cm,compat=1.17}
\begin{subfigure}{1\textwidth} 
\centering
\ref{name}
\end{subfigure}
\caption{Ablation study at a different number of generated CF (i.e. $\mid \mathcal{C}_\mathbf{x} \mid$) for each sample and with three different sensitive feature classifiers $f_s(\cdot)$ (i.e., RF, MLP, and XGB). The result refers to Adult dataset, with gender as sensitive information and Genetic as counterfactual generation strategy.
}
    \label{fig:ablationAdult}
\end{figure}
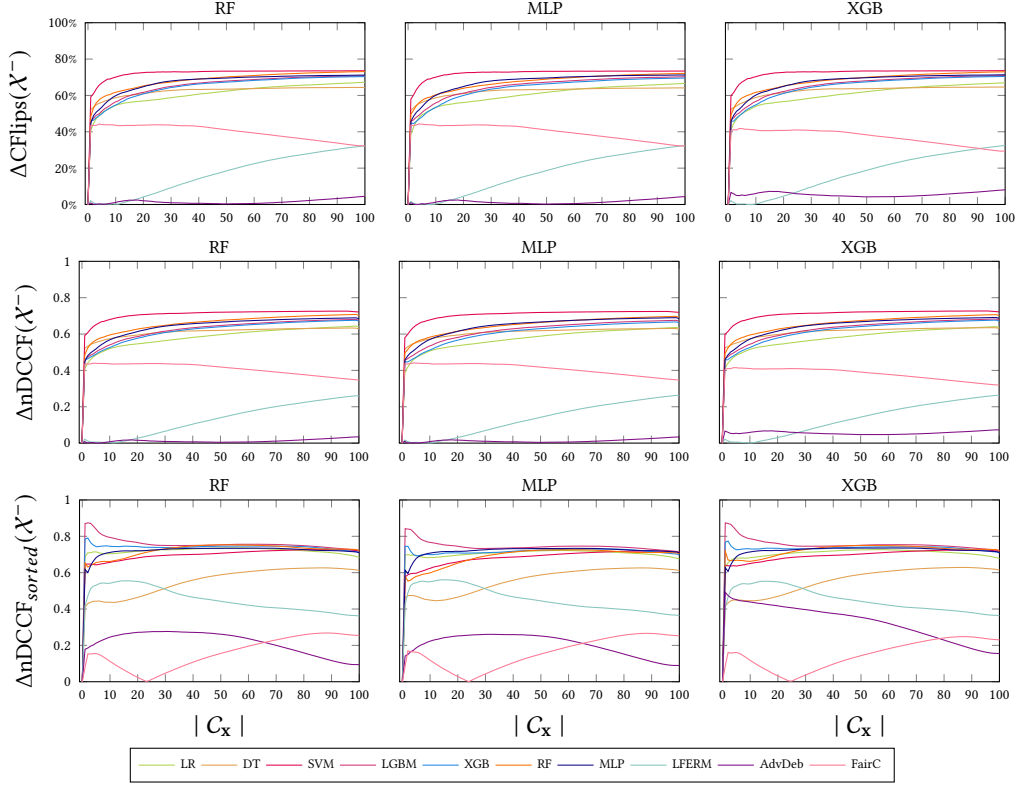

%% file: Sections/7_Conlusion.tex
In this work, we present a novel methodology for detecting bias in decision-making models that do not use sensitive features and work in a context of fairness under unawareness. Furthermore, we propose a new fairness concept (i.e., \textit{Counterfactual Fair Opportunity}) and two related fairness metrics (i.e., CFlis and nDCCF).
Understanding how an algorithm can behave with new samples and how the traits of privileged groups can influence a favorable result is of crucial importance.
Particularly, in the case of sensitive features blindness, counterfactual reasoning, and more specifically our methodology, can be an additional tool for confirming and measuring the discriminatory bias of ML models.

%% file: tables/CFstatistics.tex
\begin{table}[h]\centering
\caption{Some statistics of the three-generation strategy considered on the test set. We consider the number of samples with at least one counterfactual generated (i.e., $|\mathcal{D}|$ with $|\mathcal{C}_\mathbf{x}|>0$) and the median of the number of generated counterfactual (i.e.,$| \mathcal{C}_\mathbf{x} |$). \textbf{E.C.T.} stands for \textit{extensive computation time}, and \textbf{N.C.} stands for \textit{not compatible} model parameter.}\label{tab:StatisticsCF}
\scriptsize
\setlength{\tabcolsep}{3.5pt}
\begin{tabular}{llllrrrr}\hline\toprule
& & & &\multicolumn{3}{c}{\textbf{CF generation strategy}} \\\cmidrule(lr){5-7}
Dataset &$s$ &model &Statistics &Genetic &KDtree &MACE \\ \midrule
\multirow{8}{*}{Adult} &\multirow{8}{*}{gender} &\multirow{2}{*}{LR} &$|\mathcal{D}|$ with $|\mathcal{C}_\mathbf{x}|>0$&4523 &4523 &2854 \\
& & &median($| \mathcal{C}_\mathbf{x} |$) &100 &100 &8 \\\cmidrule(lr){3-7}
& &\multirow{2}{*}{DT} &$|\mathcal{D}|$ with $|\mathcal{C}_\mathbf{x}|>0$&4523 &4523 &3857 \\
& & &median($| \mathcal{C}_\mathbf{x} |$) &100 &100 &7 \\\cmidrule(lr){3-7}
& &\multirow{2}{*}{RF} &$|\mathcal{D}|$ with $|\mathcal{C}_\mathbf{x}|>0$&4523 &4523 &\textbf{E.C.T.} \\
& & &median($| \mathcal{C}_\mathbf{x} |$) &100 &100 &\textbf{E.C.T.} \\\cmidrule(lr){3-7}
& &\multirow{2}{*}{MLP} &$|\mathcal{D}|$ with $|\mathcal{C}_\mathbf{x}|>0$&4523 &4523 &\textbf{N.C.} \\
& & &median($| \mathcal{C}_\mathbf{x} |$) &100 &100 &\textbf{N.C.} \\ \midrule
\multirow{8}{*}{Adult-deb.} &\multirow{8}{*}{gender} &\multirow{2}{*}{LR} &$|\mathcal{D}|$ with $|\mathcal{C}_\mathbf{x}|>0$&4523 &4523 &2857 \\
& & &median($| \mathcal{C}_\mathbf{x} |$) &100 &95 &5 \\\cmidrule(lr){3-7}
& &\multirow{2}{*}{DT} &$|\mathcal{D}|$ with $|\mathcal{C}_\mathbf{x}|>0$&4523 &4523 &4183 \\
& & &median($| \mathcal{C}_\mathbf{x} |$) &98 &100 &5 \\\cmidrule(lr){3-7}
& &\multirow{2}{*}{RF} &$|\mathcal{D}|$ with $|\mathcal{C}_\mathbf{x}|>0$&4523 &4523 &3929 \\
& & &median($| \mathcal{C}_\mathbf{x} |$) &98 &81 &5 \\\cmidrule(lr){3-7}
& &\multirow{2}{*}{MLP} &$|\mathcal{D}|$ with $|\mathcal{C}_\mathbf{x}|>0$&4523 &4523 &\textbf{N.C.} \\
& & &median($| \mathcal{C}_\mathbf{x} |$) &98 &77 &\textbf{N.C.} \\\midrule
\multirow{8}{*}{Crime} &\multirow{8}{*}{race} &\multirow{2}{*}{LR} &$|\mathcal{D}|$ with $|\mathcal{C}_\mathbf{x}|>0$&200 &200 &0 \\
& & &median($| \mathcal{C}_\mathbf{x} |$) &100 &100 &0 \\\cmidrule(lr){3-7}
& &\multirow{2}{*}{DT} &$|\mathcal{D}|$ with $|\mathcal{C}_\mathbf{x}|>0$&200 &200 &\textbf{E.C.T.} \\
& & &median($| \mathcal{C}_\mathbf{x} |$) &100 &100 &\textbf{E.C.T.} \\\cmidrule(lr){3-7}
& &\multirow{2}{*}{RF} &$|\mathcal{D}|$ with $|\mathcal{C}_\mathbf{x}|>0$&200 &200 &\textbf{E.C.T.} \\
& & &median($| \mathcal{C}_\mathbf{x} |$) &100 &100 &\textbf{E.C.T.} \\\cmidrule(lr){3-7}
& &\multirow{2}{*}{MLP} &$|\mathcal{D}|$ with $|\mathcal{C}_\mathbf{x}|>0$&200 &200 &\textbf{N.C.} \\
& & &median($| \mathcal{C}_\mathbf{x} |$) &100 &100 &\textbf{N.C.} \\\midrule
\multirow{8}{*}{German} &\multirow{8}{*}{gender} &\multirow{2}{*}{LR} &$|\mathcal{D}|$ with $|\mathcal{C}_\mathbf{x}|>0$&100 &100 &33 \\
& & &median($| \mathcal{C}_\mathbf{x} |$) &100 &100 &7 \\\cmidrule(lr){3-7}
& &\multirow{2}{*}{DT} &$|\mathcal{D}|$ with $|\mathcal{C}_\mathbf{x}|>0$&100 &100 &19 \\
& & &median($| \mathcal{C}_\mathbf{x} |$) &100 &100 &9 \\\cmidrule(lr){3-7}
& &\multirow{2}{*}{RF} &$|\mathcal{D}|$ with $|\mathcal{C}_\mathbf{x}|>0$&100 &100 &16 \\
& & &median($| \mathcal{C}_\mathbf{x} |$) &98 &100 &9 \\\cmidrule(lr){3-7}
& &\multirow{2}{*}{MLP} &$|\mathcal{D}|$ with $|\mathcal{C}_\mathbf{x}|>0$&100 &100 &\textbf{N.C.} \\
& & &median($| \mathcal{C}_\mathbf{x} |$) &99 &100 &\textbf{N.C.} \\
\bottomrule\hline
\end{tabular}
\end{table}

%% file: tables/CFlipsGeneticNegOtherSensitive.tex
\begin{table*}[ht]
    \centering
    \scriptsize
    \caption{(Genetic strategy, and other sensitive features) \text{CFlip} and \text{nDCCF} results at different $|k|$ number of Counterfactuals for each negatively predicted Test set sample, ($0^*$ there are no negative predicted \textit{unprivileged} samples which result in no CF samples for the unprivileged group, hence \textbf{FairC} and \textbf{LFERM} results denoted with the $\dagger$ cannot be considered as the best result).}
    \label{tab:totGeneticCFLIPNDCCFother}
    \begin{adjustbox}{width=\textwidth, center}
\setlength{\tabcolsep}{2.5pt}
\renewcommand{\arraystretch}{1}
\rowcolors{12}{gray!15}{white}
\begin{tabular}{lllrrr|rrr|rrr|rrr|rrr|rrr}
\toprule
& & & \multicolumn{9}{c}{$\text{CFlips}$@$|k|$ (\%)} & \multicolumn{9}{c}{$\text{nDCCF}@|k|$}
\\ \cmidrule(lr){4-12} \cmidrule(lr){13-21}
& & &  \multicolumn{3}{c}{Privileged} & \multicolumn{3}{c}{Unprivileged} & \multicolumn{3}{c}{$\Delta\text{CFlips}\downarrow
$}&\multicolumn{3}{c}{Privileged} & \multicolumn{3}{c}{Unprivileged} & \multicolumn{3}{c}{$\Delta\text{nDCCF}\downarrow
$} \\ \cmidrule(lr){4-6} \cmidrule(lr){7-9} \cmidrule(lr){10-12} \cmidrule(lr){13-15} \cmidrule(lr){16-18} \cmidrule(lr){19-21}  
 Dataset  &$s$ & model   &   @10 &   @50 &   \multicolumn{1}{c}{@100} &   @10 &   @50 &   \multicolumn{1}{c}{@100} &   @10 &   @50 &   \multicolumn{1}{c}{@100} &   @10 & @50 & \multicolumn{1}{c}{@100} & @10 &   @50 &   \multicolumn{1}{c}{@100} & @10 & @50 &   @100 \\
 \hline
\midrule

& & \textbf{LR}      &          0     &          0.047 &           0.039 &         95.445 &         96.202 &          96.416 &        95.445 &        96.155 &         96.377 &        1      &        0.9996 &         0.9925 &        0.0491 &        0.0409 &         0.0379 &               0.9509 &               0.9587 &                0.9546 \\
 \cellcolor{white} &\cellcolor{white} & \textbf{DT}      &          3.263 &          2.483 &           3.075 &         71.636 &         83.817 &          85.698 &        68.373 &        81.334 &         82.623 &        0.9677 &        0.9736 &         0.9628 &        0.3148 &        0.1985 &         0.1691 &               0.6529 &               0.7751 &                0.7937 \\
& & \textbf{SVM}     &          0.292 &          0.813 &           0.883 &         94.109 &         95.632 &          96.184 &        93.817 &        94.819 &         95.301 &        0.997  &        0.9929 &         0.9847 &        0.0645 &        0.0487 &         0.0424 &               0.9325 &               0.9442 &                0.9423 \\
 \cellcolor{white} &\cellcolor{white} & \textbf{LGBM}     &          0.261 &          0.244 &           0.318 &         93.842 &         95.03  &          95.194 &        93.581 &        94.786 &         94.876 &        0.9976 &        0.9976 &         0.9899 &        0.068  &        0.0543 &         0.051  &               0.9296 &               0.9433 &                0.9389 \\
 & & \textbf{XGB}     &          0.274 &          0.239 &           0.323 &         94.284 &         95.138 &          95.248 &        94.01  &        94.899 &         94.925 &        0.9974 &        0.9976 &         0.9898 &        0.0635 &        0.0526 &         0.0498 &               0.9339 &               0.945  &                0.94   \\
 \cellcolor{white} &\cellcolor{white} & \textbf{RF}      &          0.174 &          0.37  &           0.488 &         91.872 &         93.373 &          93.844 &        91.698 &        93.003 &         93.356 &        0.9983 &        0.9967 &         0.9885 &        0.0861 &        0.0709 &         0.0652 &               0.9122 &               0.9258 &                0.9233 \\
 & & \textbf{MLP}     &          0.178 &          0.206 &           0.36  &         94.686 &         95.983 &          95.687 &        94.508 &        95.777 &         95.327 &        0.9984 &        0.9981 &         0.9897 &        0.0583 &        0.0444 &         0.0446 &               0.9401 &               0.9537 &                0.9451 \\
 \cellcolor{white} &\cellcolor{white} & \textbf{AdvDeb}  &         31.795 &         34.003 &          36.823 &         15.895 &         18.431 &          21.441 &        \textbf{15.9}   &        \textbf{15.572} &         \textbf{15.382} &        0.6903 &        0.6666 &         0.6382 &        0.8455 &        0.8232 &         0.7923 &               \textbf{0.1552} &               \textbf{0.1566} &                \textbf{0.1541} \\
 & & \textbf{LFERM}   &          0.602 &          0.927 &           1.164 &         83.469 &         86.653 &          88.216 &        82.867 &        85.726 &         87.052 &        0.9937 &        0.9914 &         0.9823 &        0.1756 &        0.1443 &         0.1274 &               0.8181 &               0.8471 &                0.8549 \\
\multirow{-10}{*}{\cellcolor{white}Adult} &\multirow{-10}{*}{\cellcolor{white}\textit{marital-status}}  & \textbf{FairC}   &         11.79  &         18.44  &          24.103 &         58.966 &         62.107 &          62.786 &        47.176 &        43.667 &         38.683 &        0.8917 &        0.8343 &         0.7793 &        0.4174 &        0.3879 &         0.3765 &               0.4743 &               0.4464 &                0.4028 \\ \midrule

 & & \textbf{LR}      & 18.739         & 25.534         & 29.058          & 54.322         & 56.660         & 55.573          &        35.583 &        31.126 &         26.515 & 0.8212        & 0.7628        & 0.7242         & 0.4690        & 0.4418        & 0.4435         & 0.3522               & 0.3210               & 0.2807                \\
 \cellcolor{white}  & \cellcolor{white}  & \textbf{DT}      & 3.330          & 5.467          & 6.565           & 96.110         & 95.195         & 94.848          &        92.78  &        89.728 &         88.283 & 0.9710        & 0.9509        & 0.9338         & 0.0409        & 0.0472        & 0.0497         & 0.9301               & 0.9037               & 0.8841                \\
 & & \textbf{SVM}     & 38.474         & 42.984         & 46.836          & 48.512         & 49.912         & 46.096          &        \textbf{10.038} &         \textbf{6.928} &          \textbf{0.74}  & 0.6280        & 0.5845        & 0.5465         & 0.4995        & 0.4974        & 0.5237         & \textbf{0.1285}               & \textbf{0.0871}               & \textbf{0.0228}                \\
   \cellcolor{white}  &\cellcolor{white} & \textbf{LGBM}     & 22.314         & 24.442         & 24.446          & 80.341         & 82.010         & 84.275          &        58.027 &        57.568 &         59.829 & 0.7800        & 0.7621        & 0.7544         & 0.2068        & 0.1871        & 0.1659         & 0.5732               & 0.5750               & 0.5885                \\
 & & \textbf{XGB}     & 25.300         & 28.450         & 29.407          & 74.602         & 74.460         & 76.608          &        49.302 &        46.01  &         47.201 & 0.7418        & 0.7220        & 0.7074         & 0.2623        & 0.2580        & 0.2390         & 0.4795               & 0.4640               & 0.4684                \\
 \cellcolor{white}  &  \cellcolor{white} & \textbf{RF}      & 4.621          & 5.766          & 7.210           & 88.211         & 91.004         & 90.755          &        83.59  &        85.238 &         83.545 & 0.9575        & 0.9460        & 0.9268         & 0.1320        & 0.1000        & 0.0980         & 0.8255               & 0.8460               & 0.8288                \\
 & & \textbf{MLP}     & 3.708          & 4.730          & 5.119           & 96.087         & 95.119         & 95.662          &        92.379 &        90.389 &         90.543 & 0.9642        & 0.9556        & 0.9445         & 0.0420        & 0.0467        & 0.0430         & 0.9222               & 0.9089               & 0.9015                \\
  \cellcolor{white}  & \cellcolor{white} & \textbf{AdvDeb}  & 25.456         & 26.605         & 28.640          & 43.095         & 52.343         & 55.839          &        17.639 &        25.738 &         27.199 & 0.7528        & 0.7386        & 0.7165         & 0.5849        & 0.5017        & 0.4629         & 0.1679               & 0.2369               & 0.2536                \\
 & & \textbf{LFERM}   & 67.751         & 65.942         & 65.190          & 27.575         & 31.909         & 33.595          &        40.176 &        34.033 &         31.595 & 0.3192        & 0.3355        & 0.3409         & 0.7294        & 0.6907        & 0.6695         & 0.4102               & 0.3552               & 0.3286                \\
\multirow{-10}{*}{\cellcolor{white}AdultDeb}  &\multirow{-10}{*}{\cellcolor{white}\textit{marital-status}}  & \textbf{FairC}   & $0^*$            & $0^*$            & $0^*$             & $0^*$            & $0^*$            & $0^*$             &         $0^\dagger$     &         $0^\dagger$     &          $0^\dagger$     & $0^*$           & $0^*$           & $0^*$            & $0^*$           & $0^*$           & $0^*$            & $0^\dagger$                  & $0^\dagger$                  & $0^\dagger$                   \\
\midrule

& & \textbf{LR}      & 4.286          & 6.571          & 6.774           & 65.000         & 82.000         & 82.855          &        60.714 &        75.429 &         76.081 & 0.9632        & 0.9394        & 0.9285         & 0.2971        & 0.1910        & 0.1769         & 0.6661               & 0.7484               & 0.7516                \\
  \cellcolor{white} & \cellcolor{white}& \textbf{DT}      & 10.455         & 10.091         & 9.909           & 60.000         & 71.000         & 77.000          &        49.545 &        60.909 &         67.091 & 0.8925        & 0.8971        & 0.8925         & 0.3492        & 0.2954        & 0.2474         & 0.5433               & 0.6017               & 0.6451                \\
 & & \textbf{SVM}     & 14.286         & 13.429         & 11.180          & $0^*$            & $0^*$            & $0^*$             &        14.286 &        13.429 &         11.18  & 0.8222        & 0.8515        & 0.8677         & $0^*$           & $0^*$           & $0^*$            & 0.8222               & 0.8515               & 0.8677                \\
 \cellcolor{white}&\cellcolor{white} & \textbf{LGBM}     & 10.000         & 7.556          & 7.233           & 80.000         & 74.000         & 78.000          &        70     &        66.444 &         70.767 & 0.8954        & 0.9181        & 0.9154         & 0.1428        & 0.2264        & 0.2081         & 0.7526               & 0.6917               & 0.7073                \\
& & \textbf{XGB}     & 9.286          & 8.429          & 8.214           & $0^*$            & $0^*$            & $0^*$             &         9.286 &         8.429 &          8.214 & 0.9105        & 0.9155        & 0.9096         & $0^*$           & $0^*$           & $0^*$            & 0.9105               & 0.9155               & 0.9096                \\
 \cellcolor{white}&\cellcolor{white} & \textbf{RF}      & 12.500         & 8.750          & 7.019           & $0^*$            & $0^*$            & $0^*$             &        12.5   &         8.75  &          7.019 & 0.8389        & 0.8920        & 0.9054         & $0^*$           & $0^*$           & $0^*$            & 0.8389               & 0.8920               & 0.9054                \\
 & & \textbf{MLP}     & 10.714         & 12.000         & 11.286          & 70.000         & 71.000         & 69.500          &        59.286 &        59     &         \textbf{58.214} & 0.8922        & 0.8855        & 0.8835         & 0.2960        & 0.3048        & 0.3094         & 0.5962               & 0.5807               & 0.5741                \\
 \cellcolor{white}&\cellcolor{white} & \textbf{AdvDeb}  & 7.143          & 8.000          & 7.643           & 74.000         & 77.200         & 76.200          &        66.857 &        69.2   &         68.557 & 0.9477        & 0.9299        & 0.9215         & 0.2651        & 0.2369        & 0.2400         & 0.6826               & 0.6930               & 0.6815                \\
 & & \textbf{LFERM}   & $0^*$            & $0^*$            & $0^*$             & $0^*$            & $0^*$            & $0^*$             &         $0^\dagger$     &         $0^\dagger$     &          $0^\dagger$     & $0^*$           & $0^*$           & $0^*$            & $0^*$           & $0^*$           & $0^*$            & $0^\dagger$                  & $0^\dagger$                  & $0^\dagger$                   \\
\multirow{-10}{*}{\cellcolor{white}German} &\multirow{-10}{*}{\cellcolor{white}\textit{age}} & \textbf{FairC}   & 8.462          & 10.769         & 9.629           & 40.000         & 64.000         & 69.000          &        \textbf{31.538} &        \textbf{53.231} &         59.371 & 0.9071        & 0.8934        & 0.8957         & 0.6621        & 0.4324        & 0.3572         & \textbf{0.2450}               & \textbf{0.4610}               & \textbf{0.5385}                \\
\bottomrule
\hline
\end{tabular}
\end{adjustbox}
\end{table*}

%% file: tables/CFlipsKDtreeNegOtherSensitive.tex
\begin{table*}[ht]
    \centering
    \scriptsize
    \caption{(KDtree strategy, and other sensitive features) \text{CFlip} and \text{nDCCF} results at different $|k|$ number of Counterfactuals for each negatively predicted Test set sample, ($0^*$ there are no negative predicted \textit{unprivileged} samples which result in no CF samples for the unprivileged group, hence \textbf{LFERM} results denoted with the $\dagger$ cannot be considered as the best result).}
    \label{tab:totKDtreeCFLIPNDCCFother}
    \begin{adjustbox}{width=\textwidth, center}
\setlength{\tabcolsep}{2.5pt}
\renewcommand{\arraystretch}{1}
\rowcolors{12}{gray!15}{white}
\begin{tabular}{lllrrr|rrr|rrr|rrr|rrr|rrr}
\hline
\toprule
& & & \multicolumn{9}{c}{$\text{CFlips}$@$|k|$ (\%)} & \multicolumn{9}{c}{$\text{nDCCF}@|k|$}
\\ \cmidrule(lr){4-12} \cmidrule(lr){13-21}
& & &  \multicolumn{3}{c}{Privileged} & \multicolumn{3}{c}{Unprivileged} & \multicolumn{3}{c}{$\Delta\text{CFlips}\downarrow
$}&\multicolumn{3}{c}{Privileged} & \multicolumn{3}{c}{Unprivileged} & \multicolumn{3}{c}{$\Delta\text{nDCCF}\downarrow
$} \\ \cmidrule(lr){4-6} \cmidrule(lr){7-9} \cmidrule(lr){10-12} \cmidrule(lr){13-15} \cmidrule(lr){16-18} \cmidrule(lr){19-21}  
 Dataset   &$s$ & model   &   @10 &   @50 &   \multicolumn{1}{c}{@100} &   @10 &   @50 &   \multicolumn{1}{c}{@100} &   @10 &   @50 &   \multicolumn{1}{c}{@100} &   @10 & @50 & \multicolumn{1}{c}{@100} & @10 &   @50 &   \multicolumn{1}{c}{@100} & @10 & @50 &   @100 \\
 \hline
\midrule

 & & \textbf{LR}      &          0.151 &          0.513 &           0.899 &         94.166 &         94.324 &          94.156 &        94.015 &        93.811 &         93.257 &        0.9983 &        0.9956 &         0.9855 &        0.0612 &        0.0579 &         0.0584 &               0.9371 &               0.9377 &                0.9271 \\
  \cellcolor{white}&\cellcolor{white}& \textbf{DT}      &          2.234 &          2.203 &           2.546 &         84.526 &         90.674 &          91.688 &        82.292 &        88.471 &         89.142 &        0.9775 &        0.9779 &         0.9684 &        0.1819 &        0.1151 &         0.0984 &               0.7956 &               0.8628 &                0.87   \\
  && \textbf{SVM}     &          0.981 &          0.858 &           0.906 &         93.419 &         96.441 &          96.964 &        92.438 &        95.583 &         96.058 &        0.9893 &        0.9909 &         0.9836 &        0.0828 &        0.0476 &         0.0388 &               0.9065 &               0.9433 &                0.9448 \\
  \cellcolor{white}&\cellcolor{white}& \textbf{LGBM}     &          0.212 &          0.634 &           0.945 &         95.456 &         95.195 &          95.211 &        95.244 &        94.561 &         94.266 &        0.9978 &        0.9946 &         0.9848 &        0.0467 &        0.048  &         0.0475 &               0.9511 &               0.9466 &                0.9373 \\
  && \textbf{XGB}     &          0.193 &          0.649 &           0.969 &         95.339 &         95.24  &          95.239 &        95.146 &        94.591 &         94.27  &        0.9981 &        0.9946 &         0.9847 &        0.0478 &        0.0479 &         0.0475 &               0.9503 &               0.9467 &                0.9372 \\
  \cellcolor{white}&\cellcolor{white}& \textbf{RF}      &          0.464 &          1.284 &           1.83  &         91.578 &         91.659 &          92.003 &        91.114 &        90.375 &         90.173 &        0.9958 &        0.9892 &         0.9773 &        0.0874 &        0.0846 &         0.081  &               0.9084 &               0.9046 &                0.8963 \\
  && \textbf{MLP}     &          0.147 &          0.504 &           0.861 &         96.555 &         96.068 &          96.1   &        96.408 &        95.564 &         95.239 &        0.9987 &        0.9958 &         0.9857 &        0.0343 &        0.0383 &         0.0382 &               0.9644 &               0.9575 &                0.9475 \\
  \cellcolor{white}&\cellcolor{white}& \textbf{AdvDeb}  &         48.64  &         52.045 &          52.572 &          8.673 &         14.166 &          19.243 &        \textbf{39.967} &        \textbf{37.879} &         \textbf{33.329} &        0.5171 &        0.4879 &         0.4774 &        0.9199 &        0.8733 &         0.8238 &               \textbf{0.4028} &               \textbf{0.3854} &                \textbf{0.3464} \\
  & &\textbf{LFERM}   &          0.488 &          1.331 &           2.191 &         88.894 &         90.653 &          90.784 &        88.406 &        89.322 &         88.593 &        0.9957 &        0.9888 &         0.9745 &        0.1149 &        0.0984 &         0.0948 &               0.8808 &               0.8904 &                0.8797 \\
 \multirow{-10}{*}{\cellcolor{white}Adult} &\multirow{-10}{*}{\cellcolor{white}\textit{marital-status}} & \textbf{FairC}   &          0.041 &          0.171 &           0.33  &         98.096 &         97.942 &          97.807 &        98.055 &        97.771 &         97.477 &        0.9995 &        0.9986 &         0.9901 &        0.0194 &        0.0203 &         0.0213 &               0.9801 &               0.9783 &                0.9688 \\
\midrule
& & \textbf{LR}      &         24.853 &         23.852 &          24.977 &         53.391 &         58.593 &          57.147 &        28.538 &        34.741 &         32.17  &        0.7586 &        0.7541 &         0.7404 &        0.4707 &        0.4258 &         0.4338 &               0.2879 &               0.3283 &                0.3066 \\
  \cellcolor{white}&\cellcolor{white}& \textbf{DT}      &          3.374 &          4.904 &           5.025 &         91.911 &         94.702 &          95.041 &        88.537 &        89.798 &         90.016 &        0.9702 &        0.9557 &         0.9441 &        0.0847 &        0.0592 &         0.054  &               0.8855 &               0.8965 &                0.8901 \\
  & & \textbf{SVM}     &         24.813 &         27.264 &          28.449 &         51.169 &         52.049 &          50.253 &        \textbf{26.356} &        \textbf{24.785} &         \textbf{21.804} &        0.7555 &        0.7273 &         0.714  &        0.4827 &        0.4717 &         0.4829 &               \textbf{0.2728} &               \textbf{0.2556} &                \textbf{0.2311} \\
  \cellcolor{white}&\cellcolor{white}& \textbf{LGBM}     &          4.535 &          2.223 &           2.255 &         94.597 &         96.868 &          97.2   &        90.062 &        94.645 &         94.945 &        0.9594 &        0.9741 &         0.9667 &        0.0554 &        0.0366 &         0.032  &               0.904  &               0.9375 &                0.9347 \\
 & & \textbf{XGB}     &          1.054 &          1.291 &           1.752 &         98.578 &         98.585 &          98.101 &        97.524 &        97.294 &         96.349 &        0.9894 &        0.9867 &         0.9747 &        0.0147 &        0.0146 &         0.018  &               0.9747 &               0.9721 &                0.9567 \\
  \cellcolor{white}&\cellcolor{white}& \textbf{RF}      &          5.01  &          8.67  &           8.846 &         85.992 &         86.652 &          87.758 &        80.982 &        77.982 &         78.912 &        0.9555 &        0.9212 &         0.9101 &        0.1336 &        0.1341 &         0.1243 &               0.8219 &               0.7871 &                0.7858 \\
 & & \textbf{MLP}     &          3.746 &          4.894 &           5.137 &         93.47  &         94.774 &          95.116 &        89.724 &        89.88  &         89.979 &        0.9706 &        0.9562 &         0.9434 &        0.0602 &        0.0546 &         0.0509 &               0.9104 &               0.9016 &                0.8925 \\
  \cellcolor{white}&\cellcolor{white}& \textbf{AdvDeb}  &          1.729 &          3.122 &           3.671 &         95.648 &         97.167 &          97.588 &        93.919 &        94.045 &         93.917 &        0.9849 &        0.9726 &         0.9577 &        0.0464 &        0.0327 &         0.0278 &               0.9385 &               0.9399 &                0.9299 \\
  & & \textbf{LFERM}   &         77.834 &         68.732 &          66.834 &         18.86  &         26.796 &          32.561 &        58.974 &        41.936 &         34.273 &        0.2049 &        0.2858 &         0.3083 &        0.8232 &        0.7533 &         0.6968 &               0.6183 &               0.4675 &                0.3885 \\
 \multirow{-10}{*}{\cellcolor{white}AdultDeb} &\multirow{-10}{*}{\cellcolor{white}\textit{marital-status}} & \textbf{FairC}   &          1.387 &          2.916 &           3.385 &         95.893 &         96.981 &          97.426 &        94.506 &        94.065 &         94.041 &        0.9878 &        0.9748 &         0.9605 &        0.0438 &        0.0337 &         0.0289 &               0.944  &               0.9411 &                0.9316 \\
\midrule
 & & \textbf{LR}      & 10.952         & 13.048         & 13.000          & 62.500         & 75.000         & 75.750          &        51.548 &        61.952 &         62.75  & 0.8854        & 0.8733        & 0.8670         & 0.3522        & 0.2719        & 0.2588         & 0.5332               & 0.6014               & 0.6082                \\
  \cellcolor{white}&\cellcolor{white}& \textbf{DT}      & 15.909         & 16.182         & 15.909          & 50.000         & 67.000         & 74.500          &        34.091 &        50.818 &         58.591 & 0.8222        & 0.8320        & 0.8299         & 0.4871        & 0.3649        & 0.2898         & 0.3351               & 0.4671               & 0.5401                \\
  & & \textbf{SVM}     & 8.571          & 12.571         & 16.000          & $0^*$            & $0^*$            & $0^*$             &         8.571 &        12.571 &         16     & 0.8858        & 0.8733        & 0.8414         & $0^*$           & $0^*$           & $0^*$            & 0.8858               & 0.8733               & 0.8414                \\
  \cellcolor{white}&\cellcolor{white}& \textbf{LGBM}     & 10.000         & 10.889         & 14.889          & 80.000         & 72.000         & 76.000          &        70     &        61.111 &         61.111 & 0.8866        & 0.8878        & 0.8527         & 0.1642        & 0.2552        & 0.2344         & 0.7224               & 0.6326               & 0.6183                \\
  && \textbf{XGB}     & 10.000         & 11.714         & 14.357          & $0^*$            & $0^*$            & $0^*$             &        10     &        11.714 &         14.357 & 0.8949        & 0.8854        & 0.8574         & $0^*$           & $0^*$           & $0^*$            & 0.8949               & 0.8854               & 0.8574                \\
  \cellcolor{white}&\cellcolor{white}& \textbf{RF}      & 10.000         & 12.750         & 15.125          & $0^*$            & $0^*$            & $0^*$             &        10     &        12.75  &         15.125 & 0.8773        & 0.8687        & 0.8458         & $0^*$           & $0^*$           & $0^*$            & 0.8773               & 0.8687               & 0.8458                \\
  && \textbf{MLP}     & 10.000         & 14.000         & 13.857          & 55.000         & 59.000         & 67.500          &        45     &        45     &         53.643 & 0.8965        & 0.8672        & 0.8589         & 0.4226        & 0.4137        & 0.3441         & 0.4739               & 0.4535               & 0.5148                \\
  \cellcolor{white}&\cellcolor{white}& \textbf{AdvDeb}  & 6.429          & 11.143         & 12.714          & 68.000         & 68.800         & 74.400          &        61.571 &        57.657 &         61.686 & 0.9463        & 0.8994        & 0.8759         & 0.2823        & 0.3027        & 0.2598         & 0.6640               & 0.5967               & 0.6161                \\
  & &\textbf{LFERM}   & $0^*$            & $0^*$            & $0^*$             & $0^*$            & $0^*$            & $0^*$             &         $0^\dagger$     &         $0^\dagger$     &          $0^\dagger$     & $0^*$           & $0^*$           & $0^*$            & $0^*$           & $0^*$           & $0^*$            & $0^\dagger$                  & $0^\dagger$                  & $0^\dagger$                   \\
\multirow{-10}{*}{\cellcolor{white} German}  &\multirow{-10}{*}{\cellcolor{white} \textit{age}}  & \textbf{FairC}   & 13.077         & 13.538         & 16.231          & 30.000         & 56.000         & 65.000          &        \textbf{16.923} &        \textbf{42.462} &         \textbf{48.769} & 0.8592        & 0.8641        & 0.8373         & 0.6442        & 0.4743        & 0.3920         & \textbf{0.2150}               & \textbf{0.3898}               & \textbf{0.4453}                \\

\hline
\bottomrule
\hline
\end{tabular}
\end{adjustbox}
\end{table*}

%% file: tables/DeltaNDCCFRerank.tex
\begin{table*}[ht]
    \centering
    \scriptsize
    \caption{$\mathrm{nDCCF}_{sorted}$ results at different $|k|$ number of Counterfactuals and different $f_s(\cdot)$ for each negatively predicted Test set sample with both Genetic and KDtree strategy.}
    \label{tab:totReRankNDCCF}
    \begin{adjustbox}{width=\textwidth, center}
\setlength{\tabcolsep}{2.5pt}
\renewcommand{\arraystretch}{1}
\rowcolors{15}{gray!15}{white}
\begin{tabular}{lllrrr|rrr|rrr|rrr|rrr|rrr}
\hline
\toprule
& & & \multicolumn{18}{c}{$\Delta\mathrm{nDCCF}_{sorted}@|k| \downarrow$} \\
\cmidrule(lr){4-21}
& & & \multicolumn{9}{c}{Genetic} & \multicolumn{9}{c}{KDtree}
\\ \cmidrule(lr){4-12} \cmidrule(lr){13-21}
& & &  \multicolumn{3}{c}{\textbf{RF}} & \multicolumn{3}{c}{\textbf{MLP}} & \multicolumn{3}{c}{\textbf{XGB}}&\multicolumn{3}{c}{\textbf{RF}} & \multicolumn{3}{c}{\textbf{MLP}} & \multicolumn{3}{c}{\textbf{XGB}} \\ \cmidrule(lr){4-6} \cmidrule(lr){7-9} \cmidrule(lr){10-12} \cmidrule(lr){13-15} \cmidrule(lr){16-18} \cmidrule(lr){19-21}  
 Dataset &$s$  & model   &   @10 &   @50 &   \multicolumn{1}{c}{@100} &   @10 &   @50 &   \multicolumn{1}{c}{@100} &   @10 &   @50 &   \multicolumn{1}{c}{@100} &   @10 & @50 & \multicolumn{1}{c}{@100} & @10 &   @50 &   \multicolumn{1}{c}{@100} & @10 & @50 &   @100 \\
\midrule
                                                       &   & \textbf{LR}       &     0.7046 &     0.7355 &     0.6841 &       0.6837 &      0.7205 &       0.6751     & 0.6826 & 0.7254 & 0.6775  & 0.7287 &  0.7542 &   0.7512&   0.7139 &   0.7413 &   0.7421 &      0.7134 &    0.7482 &      0.7484 \\
                                        \cellcolor{white} &\cellcolor{white} & \textbf{DT}       &     0.4373 &     0.5881 &     0.6115 &       0.4493 &      0.5873 &       0.611      & 0.4455 & 0.5906 & 0.614   & 0.4684 &  0.6704 &   0.7233&   0.475  &   0.6714 &   0.7246 &      0.4638 &    0.6666 &      0.7212 \\
                                                         & & \textbf{SVM}      &     0.6565 &     0.7069 &     0.7197 &       0.6236 &      0.6949 &       0.7128     & 0.6502 & 0.7028 & 0.7183  & 0.6291 &  0.7263 &   0.7607&   0.5986 &   0.7163 &   0.7547 &      0.6187 &    0.7233 &      0.7593 \\
                                        \cellcolor{white} &\cellcolor{white} & \textbf{LGBM}      &     0.7938 &     0.7533 &     0.7198 &       0.7665 &      0.7413 &       0.7122     & 0.7863 & 0.7506 & 0.7197  & 0.7964 &  0.7904 &   0.7808&   0.7691 &   0.774  &   0.7697 &      0.7978 &    0.7893 &      0.7815 \\
                                                          & & \textbf{XGB}      &     0.7451 &     0.7455 &     0.7105 &       0.6994 &      0.7247 &       0.6975     & 0.7305 & 0.7401 & 0.7087  & 0.8006 &  0.7971 &   0.7836&   0.7662 &   0.7769 &   0.7687 &      0.7896 &    0.793  &      0.781  \\
                                        \cellcolor{white} &\cellcolor{white} & \textbf{RF}       &     0.6617 &     0.753  &     0.7226 &       0.5988 &      0.724  &       0.7041     & 0.6624 & 0.7495 & 0.7217  & 0.6608 &  0.7285 &   0.7461&   0.5932 &   0.6935 &   0.7241 &      0.6633 &    0.7254 &      0.7441 \\
                                                          & & \textbf{MLP}      &     0.7114 &     0.7339 &     0.7095 &       0.7075 &      0.7327 &       0.7092     & 0.714 & 0.7352 & 0.7109  & 0.7645 &  0.7833 &   0.7709&   0.7566 &   0.7792 &   0.769 &      0.7664 &    0.7833 &      0.7719 \\
                                        \cellcolor{white} &\cellcolor{white} & \textbf{AdvDeb}   &     0.2457 &     0.2634 &     0.0941 &       0.2239 &      0.2536 &       0.0895     & 0.4391 & 0.3564 & 0.1553  & 0.7507 &  0.6615 &   0.5793&   0.7373 &   0.6574 &   0.578 &      0.7447 &    0.6614 &      0.5804 \\
                                                          & & \textbf{LFERM}    &     0.5421 &     0.4425 &     0.3612 &       0.5501 &      0.4446 &       0.3638     & 0.5384 & 0.4423 & 0.3627  & 0.7255 &  0.7048 &   0.6888&   0.7054 &   0.6956 &   0.6832 &      0.7239 &    0.7049 &      0.6894 \\
              \multirow{-10}{*}{\cellcolor{white}Adult}   &\multirow{-10}{*}{\cellcolor{white}\textit{gender}}   & \textbf{FairC}    &     0.1115 &     0.1559 &     0.2546 &       0.1172 &      0.1527 &       0.2531     & 0.1166 & 0.1437 & 0.2309  & 0.6931 &  0.6598 &   0.4451&   0.6818 &   0.6551 &   0.4428 &      0.7148 &    0.6691 &      0.4522 \\ \midrule
                                            & & \textbf{LR}     &    0.7997 &    0.8363 &   0.8412& 0.7503 &   0.7649 &   0.8061&    0.2753 &   0.3692 &  0.4013& 0.7617 &  0.8123 & 0.8497& 0.6112 &  0.6751 & 0.7558& 0.2279 &  0.2505 &  0.241  \\
                           \cellcolor{white} &\cellcolor{white} & \textbf{DT}     &    0.8183 &    0.7617 &   0.7526& 0.7839 &   0.7004 &   0.6973&    0.2994 &   0.3093 &  0.3519& 0.9984 &  0.9921 & 0.9854& 0.9401 &  0.9317 & 0.9123& 0.124  &  0.0448 &  0.1166 \\
                                            & & \textbf{SVM}    &    0.7578 &    0.7484 &   0.7103& 0.8609 &   0.8656 &   0.8748&    0.1569 &   0.1524 &  0.1723& 0.8384 &  0.7836 & 0.8111& 0.6282 &  0.5369 & 0.5875& 0.2841 &  0.2533 &  0.2501 \\
                           \cellcolor{white} &\cellcolor{white} & \textbf{LGBM}    &    0.9567 &    0.9297 &   0.9143& 0.9495 &   0.9383 &   0.9325&    0.7285 &   0.6479 &  0.6217& 0.9774 &  0.9855 & 0.9817& 0.9771 &  0.9799 & 0.9777& 0.2739 &  0.4302 &  0.5458 \\
                                            & & \textbf{XGB}    &    0.9884 &    0.9874 &   0.983 & 0.9752 &   0.965  &   0.9642&    0.9253 &   0.8501 &  0.7958& 0.9908 &  0.99   & 0.9849& 0.978  &  0.9684 & 0.9686& 0.8066 &  0.6459 &  0.6487 \\
                           \cellcolor{white} &\cellcolor{white} & \textbf{RF}     &    0.9965 &    0.9911 &   0.9835& 0.561  &   0.8238 &   0.8795&    0.2967 &   0.55   &  0.6561& 0.9979 &  0.9728 & 0.9716& 0.6813 &  0.8484 & 0.8839& 0.3101 &  0.4882 &  0.5277 \\
                                             & & \textbf{MLP}    &    0.9867 &    0.9942 &   0.9883& 0.9856 &   0.9941 &   0.9883&    0.9368 &   0.9252 &  0.9334& 0.9963 &  0.9981 & 0.9896& 0.9973 &  0.9985 & 0.9899& 0.5745 &  0.7063 &  0.771  \\
                           \cellcolor{white} &\cellcolor{white} & \textbf{AdvDeb} &    0.1265 &    0.3418 &   0.5543& 0.2892 &   0.1285 &   0.4043&    0.2876 &   0.0692 &  0.2024& 0.9926 &  0.9966 & 0.9887& 0.9977 &  0.9979 & 0.9895& 0.695  &  0.5866 &  0.6319 \\
                                             &&  \textbf{LFERM}  &    0.4163 &    0.5331 &   0.525 & 0.2746 &   0.4539 &   0.4738&    0.3134 &   0.3716 &  0.3764& 0.8636 &  0.7791 & 0.7236& 0.7731 &  0.7523 & 0.7132& 0.8633 &  0.7041 &  0.6417 \\
 \multirow{-10}{*}{\cellcolor{white}AdultDeb}&\multirow{-10}{*}{\cellcolor{white}\textit{gender}}& \textbf{FairC}  &    0.7758 &    0.8264 &   0.8458& 0.8373 &   0.8936 &   0.8873&    0.7472 &   0.8005 &  0.8161& 0.9872 &  0.9368 & 0.953& 0.9898 &  0.9683 & 0.9728& 0.1821 &  0.5345 &  0.6369 \\ \midrule
                                                          & & \textbf{LR}& 0.7342 & 0.7354 & 0.7828      & 0.7196 & 0.7564 & 0.7988&  0.762  & 0.7523 & 0.7962& 0.7119 & 0.7067 & 0.7606& 0.7477 & 0.7763 & 0.8259& 0.7408 & 0.7175 & 0.7671 \\
                                        \cellcolor{white} &\cellcolor{white} & \textbf{DT}& 0.3164 & 0.5576 & 0.6573      & 0.3902 & 0.6274 & 0.6886&  0.3142 & 0.5638 & 0.6609& 0.2511 & 0.4858 & 0.5909& 0.3159 & 0.5519 & 0.6497& 0.2675 & 0.4903 & 0.5948 \\
                                                         & & \textbf{SVM}& 0.5991 & 0.6745 & 0.7466     & 0.5929 & 0.7351 & 0.7917&  0.592  & 0.6746 & 0.7468& 0.443  & 0.5845 & 0.6914& 0.4874 & 0.6894 & 0.7726& 0.42   & 0.5803 & 0.688  \\
                                        \cellcolor{white} &\cellcolor{white} & \textbf{LGBM}& 0.5503 & 0.6523 & 0.7348     & 0.596  & 0.7102 & 0.775&  0.5414 & 0.6536 & 0.7349& 0.3559 & 0.5662 & 0.6863& 0.4328 & 0.6667 & 0.7708& 0.351  & 0.5646 & 0.6844 \\
                                                         & & \textbf{XGB}& 0.5414 & 0.6502 & 0.7352     & 0.6459 & 0.7117 & 0.7749&  0.5521 & 0.6571 & 0.7371& 0.5199 & 0.6134 & 0.7122& 0.6413 & 0.7318 & 0.8055& 0.5267 & 0.6159 & 0.7128 \\
                                        \cellcolor{white} &\cellcolor{white} & \textbf{RF}& 0.3876 & 0.5335 & 0.6388      & 0.3237 & 0.5068 & 0.6483&  0.4098 & 0.5398 & 0.6417& 0.1543 & 0.4155 & 0.5813& 0.252  & 0.4792 & 0.6464& 0.169  & 0.4205 & 0.584  \\
                                                         & & \textbf{MLP}& 0.4726 & 0.6319 & 0.7064     & 0.6346 & 0.7333 & 0.7816&  0.4659 & 0.6282 & 0.7015& 0.398  & 0.5685 & 0.6803& 0.4528 & 0.6601 & 0.7567& 0.4415 & 0.5852 & 0.6905 \\
                                       \cellcolor{white} & \cellcolor{white} & \textbf{AdvDeb}& 0.5562 & 0.6698 & 0.7072  & 0.5355 & 0.61   & 0.6922&  0.5443 & 0.6592 & 0.7007& 0.424  & 0.6333 & 0.6821& 0.4894 & 0.7073 & 0.7565& 0.4262 & 0.6353 & 0.6827 \\
                                                         & & \textbf{LFERM}& 0.5352 & 0.63   & 0.6554   & 0.5413 & 0.6462 & 0.6812&  0.5403 & 0.6378 & 0.6592& 0.5028 & 0.5753 & 0.6113& 0.6033 & 0.6384 & 0.686& 0.4935 & 0.5762 & 0.6092 \\
              \multirow{-10}{*}{\cellcolor{white}Crime}&\multirow{-10}{*}{\cellcolor{white}\textit{race}}& \textbf{FairC}& 0.2352 & 0.2452 & 0.2575   & 0.2605 & 0.2871 & 0.309&  0.2381 & 0.2554 & 0.2655& 0.3127 & 0.1832 & 0.1983& 0.3177 & 0.231  & 0.2616& 0.3221 & 0.1919 & 0.2067 \\ \midrule
                                                       &   & \textbf{LR}    &       0.2684 &  0.3386 &  0.3711  &  1 &     1 &    0.9928& 0.2569 &  0.1951 &  0.1426&   0.8209 &  0.8272 & 0.7799 &1 &  1 &  0.9928 &  0.2988 &  0.351  &  0.3507 \\
                          \cellcolor{white} &\cellcolor{white} & \textbf{DT}    &       0.8207 &  0.8973 &  0.9038  &  1 &     1 &    0.9928& 0.0456 &  0.1073 &  0.1735&   0.8102 &  0.8682 & 0.8887 &1 &  1 &  0.9928 &  0.0793 &  0.1999 &  0.2701 \\
                                           & & \textbf{SVM}   &       0.8709 &  0.9048 &  0.9208  &  1 &     1 &    0.9928& 0.0487 &  0.2179 &  0.3001&   0.8014 &  0.8704 & 0.895  &1 &  1 &  0.9928 &  0.0608 &  0.2932 &  0.3328 \\
                          \cellcolor{white} &\cellcolor{white} & \textbf{LGBM}   &       0.1931 &  0.1601 &  0.1673  &  1 &     1 &    0.9928& 0.2344 &  0.1529 &  0.1408&   0.6436 &  0.6872 & 0.6971 &1 &  1 &  0.9928 &  0.2972 &  0.2794 &  0.2855 \\
                                           & & \textbf{XGB}   &       0.1843 &  0.0727 &  0.1543  &  1 &     1 &    0.9928& 0.2158 &  0.1961 &  0.114&    0.628  &  0.7153 & 0.7031 &1 &  1 &  0.9928 &  0.2278 &  0.3131 &  0.3076 \\
                          \cellcolor{white} &\cellcolor{white} & \textbf{RF}    &       0.9751 &  0.9555 &  0.9531  &  1 &     1 &    0.9927& 0.0182 &  0.0425 &  0.0796&   0.8537 &  0.8898 & 0.9093 &1 &  1 &  0.9928 &  0.4025 &  0.3219 &  0.3631 \\
                                            & & \textbf{MLP}   &       0.8684 &  0.9149 &  0.9226  &  1 &     1 &    0.9928& 0.5936 &  0.6574 &  0.677 &   0.7799 &  0.8557 & 0.8812 &1 &  1 &  0.9928 &  0.4797 &  0.6189 &  0.6581 \\
                          \cellcolor{white} &\cellcolor{white} & \textbf{AdvDeb}&       0.0203 &  0.0249 &  0.0681  &  1 &     1 &    0.9928& 0.2212 &  0.1629 &  0.1276&   0.7353 &  0.7113 & 0.7032 &1 &  1 &  0.9928 &  0.3325 &  0.2851 &  0.3151 \\
                                           & & \textbf{LFERM} &       0.89   &  0.9204 &  0.9165  &  1 &     1 &    0.9928& 0.7682 &  0.8319 &  0.7623&   1      &  0.9773 & 0.9676 &1 &  1 &  0.9928 &  0.6504 &  0.7021 &  0.7281 \\
\multirow{-10}{*}{\cellcolor{white}German}&\multirow{-10}{*}{\cellcolor{white}\textit{gender}}& \textbf{FairC} &       0.9782 &  0.9483 &  0.9435  &  1 &     1 &    0.9928& 0.3023 &  0.2761 &  0.2498&   0.9144 &  0.8805 & 0.889  &1 &  1 &  0.9928 &  0.0652 &  0.1668 &  0.253  \\
\bottomrule
\hline
\end{tabular}
\end{adjustbox}
\end{table*}

%% file: tables/DeltaNDCCFRerankOtherSensitive.tex
\begin{table*}[ht]
    \centering
    \scriptsize
    \caption{$\mathrm{nDCCF}_{sorted}$ results at different $|k|$ number of Counterfactuals and different $f_s(\cdot)$ for each negatively predicted Test set sample with both Genetic and KDtree strategy and other sensitive features. Results denoted with $\dagger$ cannot be considered best model results since there are no negative predicted samples with the sensitive information correctly predicted.}
    \label{tab:totReRanknDCCFothersensitive}
    \begin{adjustbox}{width=\textwidth, center}
\setlength{\tabcolsep}{2.5pt}
\renewcommand{\arraystretch}{1}
\rowcolors{15}{gray!15}{white}
\begin{tabular}{lllrrr|rrr|rrr|rrr|rrr|rrr}
\hline
\toprule
& & & \multicolumn{18}{c}{$\Delta\mathrm{nDCCF}_{sorted}@|k| \downarrow$} \\
\cmidrule(lr){4-21}
& & & \multicolumn{9}{c}{Genetic} & \multicolumn{9}{c}{KDtree}
\\ \cmidrule(lr){4-12} \cmidrule(lr){13-21}
& & &  \multicolumn{3}{c}{\textbf{RF}} & \multicolumn{3}{c}{\textbf{MLP}} & \multicolumn{3}{c}{\textbf{XGB}}&\multicolumn{3}{c}{\textbf{RF}} & \multicolumn{3}{c}{\textbf{MLP}} & \multicolumn{3}{c}{\textbf{XGB}} \\ \cmidrule(lr){4-6} \cmidrule(lr){7-9} \cmidrule(lr){10-12} \cmidrule(lr){13-15} \cmidrule(lr){16-18} \cmidrule(lr){19-21}  
 Dataset &$s$  & model   &   @10 &   @50 &   \multicolumn{1}{c}{@100} &   @10 &   @50 &   \multicolumn{1}{c}{@100} &   @10 &   @50 &   \multicolumn{1}{c}{@100} &   @10 & @50 & \multicolumn{1}{c}{@100} & @10 &   @50 &   \multicolumn{1}{c}{@100} & @10 & @50 &   @100 \\

\midrule
                                              &                                                   & \textbf{LR}    & 0.8469 & 0.9179 & 0.9364&  0.849  &  0.9203 &  0.9393& 0.8491 & 0.9203 & 0.9392& 0.817  & 0.8736 & 0.905  & 0.815  & 0.8723 & 0.9041& 0.8163 & 0.873  & 0.9047 \\
                            \cellcolor{white} &                                  \cellcolor{white}& \textbf{DT}    & 0.2218 & 0.6785 & 0.7467&  0.195  &  0.6689 &  0.7395& 0.1962 & 0.6692 & 0.7395& 0.0684 & 0.6631 & 0.769  & 0.0613 & 0.6599 & 0.7659& 0.0626 & 0.6603 & 0.766  \\
                                              &                                                   & \textbf{SVM}   & 0.7871 & 0.9072 & 0.9203&  0.7887 &  0.9085 &  0.9212& 0.7892 & 0.9089 & 0.9214& 0.5777 & 0.845  & 0.8925 & 0.5751 & 0.8441 & 0.8914& 0.5741 & 0.8438 & 0.8913 \\
                            \cellcolor{white} &                                  \cellcolor{white}& \textbf{LGBM}   & 0.9476 & 0.9671 & 0.9465&  0.9466 &  0.9665 &  0.9457& 0.9472 & 0.9671 & 0.9461& 0.9298 & 0.9352 & 0.9339 & 0.928  & 0.9344 & 0.933 & 0.9283 & 0.9346 & 0.9333 \\
                                              &                                                   & \textbf{XGB}   & 0.9489 & 0.9683 & 0.9477&  0.9485 &  0.9681 &  0.9473& 0.9489 & 0.9682 & 0.9474& 0.9296 & 0.9337 & 0.9337 & 0.9286 & 0.9326 & 0.9328& 0.9287 & 0.9328 & 0.933  \\
                            \cellcolor{white} &                                  \cellcolor{white}& \textbf{RF}    & 0.6507 & 0.844  & 0.8845&  0.6487 &  0.843  &  0.8838& 0.65   & 0.8434 & 0.8841& 0.5529 & 0.774  & 0.8396 & 0.5484 & 0.7716 & 0.8378& 0.5493 & 0.7722 & 0.8382 \\
                                              &                                                   & \textbf{MLP}   & 0.902  & 0.9547 & 0.9434&  0.905  &  0.956  &  0.9442& 0.9046 & 0.9558 & 0.9441& 0.7818 & 0.898  & 0.9188 & 0.7812 & 0.8976 & 0.9185& 0.782  & 0.898  & 0.9188 \\
                            \cellcolor{white} &                                  \cellcolor{white}& \textbf{AdvDeb}& 0.2361 & 0.321  & 0.1938&  0.2206 &  0.3076 &  0.1855& 0.2211 & 0.3087 & 0.1862& 0.8374 & 0.7608 & 0.4466 & 0.8382 & 0.7635 & 0.4503& 0.8382 & 0.7641 & 0.4505 \\
                                              &                                                   & \textbf{LFERM} & 0.8654 & 0.8961 & 0.87  &  0.8739 &  0.8998 &  0.8721& 0.8746 & 0.9    & 0.8722& 0.8001 & 0.8467 & 0.8649 & 0.7998 & 0.8454 & 0.8638& 0.8    & 0.8456 & 0.8639 \\
  \multirow{-10}{*}{\cellcolor{white}Adult}   & \multirow{-10}{*}{\cellcolor{white}\textit{marital-status}}& \textbf{FairC} & 0.4182 & 0.4114 & 0.3427&  0.5743 &  0.589  &  0.5538& 0.4913 & 0.4854 & 0.4115& 0.9791 & 0.9796 & 0.9693 & 0.9789 & 0.9794 & 0.9688& 0.9795 & 0.9797 & 0.969  \\ \midrule
                                              &                                                   & \textbf{LR}     & 0.2912   & 0.2283    & 0.3716  & 0.4241    & 0.3303    & 0.4710  & 0.3489  & 0.2526  & 0.2697& 0.1561 &  0.1121 & 0.1493& 0.1906 & 0.1994 & 0.2067& 0.3417 & 0.278 & 0.3204  \\
                          \cellcolor{white} &                                  \cellcolor{white}& \textbf{DT}    & 0.9793   & 0.9612    & 0.9446  & 0.9905    & 0.8720    & 0.7839  & 0.9069  & 0.8842  & 0.8793& 0.982  &  0.9716 & 0.9485& 1      & 0.9378 & 0.6835& 0.9074 & 0.9121 & 0.8945 \\
                                            &                                                   & \textbf{SVM}   & 0.6015   & 0.7144    & 0.7673  & 0.6397    & 0.7266    & 0.8116  & 0.1212  & 0.0940  & 0.0253& 0.0722 &  0.0595 & 0.0337& 0.2333 & 0.1626 & 0.1344& 0.0954 & 0.145 & 0.1721  \\
                          \cellcolor{white} &                                  \cellcolor{white}& \textbf{LGBM}   & 0.7785   & 0.8785    & 0.8774  & 0.7113    & 0.6058    & 0.6457  & 0.8836  & 0.8259  & 0.6625& 0.6504 &  0.8552 & 0.8965& 0.7551 & 0.5747 & 0.5786& 0.8752 & 0.9176 & 0.932  \\
                                            &                                                   & \textbf{XGB}   & 0.7300   & 0.8553    & 0.8546  & 0.4647    & 0.5820    & 0.6475  & 0.6417  & 0.6759  & 0.5171& 0.6516 &  0.8587 & 0.8971& 0.5574 & 0.4886 & 0.5438& 0.9588 & 0.9401 & 0.948  \\
                          \cellcolor{white} &                                  \cellcolor{white}& \textbf{RF}    & 0.4504   & 0.6144    & 0.7334  & 0.4139    & 0.4397    & 0.5489  & 0.4703  & 0.6615  & 0.7627& 0.4918 &  0.5385 & 0.6809& 0.5461 & 0.2726 & 0.3601& 0.6909 & 0.6395 & 0.7468 \\
                                            &                                                   & \textbf{MLP} & 0.6973   & 0.8844    & 0.9176  & 0.6835    & 0.8255    & 0.8628  & 0.6527  & 0.8218  & 0.8620& 0.5547 &  0.8305 & 0.859& 0.1866 & 0.4874 & 0.5805& 0.791 & 0.8594 & 0.8715  \\
                          \cellcolor{white} &                                  \cellcolor{white}& \textbf{AdvDeb}& 0.7540   & 0.3037    & 0.0915  & 0.9283    & 0.4493    & 0.0600  & 0.6822  & 0.2555  & 0.0919& 0.9867 &  0.9844 & 0.9747& 0.8529 & 0.6696 & 0.7141& 0.9495 & 0.9439 & 0.9314 \\
                                            &                                                   & \textbf{LFERM} & 0.5720   & 0.2884    & 0.2625  & 0.4149    & 0.4856    & 0.4693  & 0.9271  & 0.3538  & 0.3924& 0.8112 &  0.4704 & 0.2734& 0.8812 & 0.7315 & 0.6934& 0.9946 & 0.6949 & 0.4657 \\
\multirow{-10}{*}{\cellcolor{white}AdultDeb}& \multirow{-10}{*}{\cellcolor{white}\textit{marital-status}}& \textbf{FairC} & $0^\dagger$   & $0^\dagger$    & $0^\dagger$  & $0^\dagger$    & $0^\dagger$    & $0^\dagger$  & $0^\dagger$  & $0^\dagger$  & $0^\dagger$& 0.9784 &  0.9802 & 0.972& 0.7504 & 0.8266 & 0.8329& 0.944 & 0.9411 & 0.9316  \\ \midrule
                                             &                                        & \textbf{LR}    &0.7864   & 0.7907   & 0.8204 & 1.0000 & 1.0000  & 0.9928& 0.6381  & 0.6815  & 0.7356 & 0.7968 & 0.7463  & 0.7627   & 1.0000  & 1.0000  & 0.9928& 0.6083 & 0.5432  & 0.5988 \\
                           \cellcolor{white} &                       \cellcolor{white}& \textbf{DT}    &0.9569   & 0.7645   & 0.7871 & 1.0000 & 1.0000  & 0.9928& 0.7461  & 0.6439  & 0.6621 & 0.8209 & 0.5777  & 0.6824   & 1.0000  & 1.0000  & 0.9928& 0.2490 & 0.3767  & 0.5086 \\
                                             &                                        & \textbf{SVM}   &0.9265   & 0.9300   & 0.9417 & 1.0000 & 1.0000  & 0.9928& 0.9448  & 0.8801  & 0.8827 & 0.7425 & 0.8497  & 0.8699   & 1.0000  & 1.0000  & 0.9928& 0.7385 & 0.7967  & 0.8122 \\
                           \cellcolor{white} &                       \cellcolor{white}& \textbf{LGBM}   &0.4684   & 0.5509   & 0.6332 & 1.0000 & 1.0000  & 0.9928& 0.3275  & 0.5151  & 0.6276 & 0.5207 & 0.6687  & 0.6747   & 1.0000  & 1.0000  & 0.9928& 0.3802 & 0.5409  & 0.5851 \\
                                             &                                        & \textbf{XGB}   &0.9657   & 0.9610   & 0.9588 & 1.0000 & 1.0000  & 0.9928& 0.8588  & 0.9003  & 0.9032 & 0.8569 & 0.8893  & 0.9081   & 1.0000  & 1.0000  & 0.9928& 0.7603 & 0.8067  & 0.8347 \\
                           \cellcolor{white} &                       \cellcolor{white}& \textbf{RF}    &0.9327   & 0.9415   & 0.9309 & 1.0000 & 1.0000  & 0.9928& 0.9343  & 0.9443  & 0.9258 & 0.9033 & 0.8870  & 0.8995   & 1.0000  & 1.0000  & 0.9928& 0.7835 & 0.7906  & 0.8260 \\
                                             &                                        & \textbf{MLP}   &0.6030   & 0.6844   & 0.6959 & 1.0000 & 1.0000  & 0.9928& 0.1879  & 0.3839  & 0.5016 & 0.4127 & 0.5774  & 0.6115   & 1.0000  & 1.0000  & 0.9928& 0.2665 & 0.3935  & 0.4883 \\
                           \cellcolor{white} &                       \cellcolor{white}& \textbf{AdvDeb}&0.7600   & 0.7954   & 0.8305 & 1.0000 & 1.0000  & 0.9928& 0.6488  & 0.6116  & 0.6613 & 0.8100 & 0.7519  & 0.7835   & 1.0000  & 1.0000  & 0.9928& 0.7551 & 0.5997  & 0.6192 \\
                                             &                                        & \textbf{LFERM} &$0^\dagger$   & $0^\dagger$   & $0^\dagger$ & $0^\dagger$ & $0^\dagger$  & $0^\dagger$& $0^\dagger$  & $0^\dagger$  & $0^\dagger$ & $0^\dagger$ & $0^\dagger$  & $0^\dagger$   & $0^\dagger$  & $0^\dagger$  & $0^\dagger$& $0^\dagger$ & $0^\dagger$  & $0^\dagger$ \\
 \multirow{-10}{*}{\cellcolor{white}German}  & \multirow{-10}{*}{\cellcolor{white}\textit{age}}& \textbf{FairC} &0.9853   & 0.8394   & 0.8348 & 1.0000 & 1.0000  & 0.9928& 0.8314  & 0.6122  & 0.6093 & 0.8924 & 0.7065  & 0.6882   & 1.0000  & 1.0000  & 0.9928& 0.6765 & 0.4831  & 0.4915 \\
\bottomrule
\hline
\end{tabular}
\end{adjustbox}
\end{table*}